\documentclass[letterpaper, journal]{IEEEtran}
\usepackage{amsmath, amsfonts}
\usepackage{algorithmic}
\usepackage{algorithm}
\usepackage{array}
\usepackage[caption=false,font=normalsize,labelfont=sf,textfont=sf]{subfig}
\usepackage{textcomp}
\usepackage{stfloats}
\usepackage{url}
\usepackage{verbatim}
\usepackage{graphicx}
\usepackage{cite}
\usepackage{colortbl}
\usepackage{multirow}
\usepackage[table]{xcolor}
\usepackage{booktabs}
\usepackage{mathrsfs}
\usepackage{diagbox}
\usepackage{amssymb}

\usepackage{hyperref}
\hypersetup{
	colorlinks=true, linkcolor=blue, citecolor=green, urlcolor=magenta}

\begin{document}

\title{vSTMD: Visual Motion Detection for Extremely Tiny Target at Various Velocities}

\author{
	Mingshuo~Xu\textsuperscript{\textdagger}, 
	Hao~Luan\textsuperscript{\textdagger},
	Zhou~Daniel~Hao,~\IEEEmembership{Member,~IEEE},
	Jigen~Peng, and 
	Shigang~Yue\textsuperscript{\textasteriskcentered},~\IEEEmembership{Senior~Member,~IEEE}
	
	\thanks{This work has been submitted to the IEEE for possible publication. Copyright may be transferred without notice, after which this version may no longer be accessible.}
} 



\maketitle

\begin{abstract}
	Visual motion detection for extremely tiny (ET-) targets is challenging, due to their category-independent nature and the scarcity of visual cues, which often incapacitate mainstream feature-based models. Natural architectures with rich interpretability offer a promising alternative, where STMD architectures derived from insect visual STMD (Small Target Motion Detector) pathways have demonstrated their effectiveness. However, previous STMD models are constrained to a narrow velocity range, hindering their efficacy in real-world scenarios where targets exhibit diverse and unstable dynamics. To address this limitation, we present vSTMD, a learning-free model for motion detection of ET-targets at various velocities. Our key innovations include: (1) a cross-Inhibition Dynamic Potential (cIDP) that serves as a self-adaptive mechanism efficiently capturing motion cues across a wide velocity spectrum, and (2) the first Collaborative Directional Gradient Calculation (CDGC) strategy, which enhances orienting accuracy and robustness while reducing computational overhead to one-eighth of previously isolated strategies. Evaluated on the real-world dataset RIST, the proposed vSTMD and its feedback-facilitated variant vSTMD-F achieve relative $F_{1}$ gains of $30\%$ and $58\%$ over state-of-the-art (SOTA) STMD approaches, respectively. Furthermore, both models demonstrate competitive orientation estimation performance compared to SOTA deep learning-driven methods. Experiments also reveal the superiority of the natural architecture for ET-object motion detection - vSTMD is $60\times$ faster than contemporary data-driven methods, making it highly suitable for real-time applications in dynamic scenarios and complex backgrounds. Code is available at \href{https://github.com/MingshuoXu/vSTMD}{Project Repository}.
\end{abstract}


\section{Introduction}
\begin{figure}[!t]
	\centering
	\includegraphics[width=0.5\textwidth]{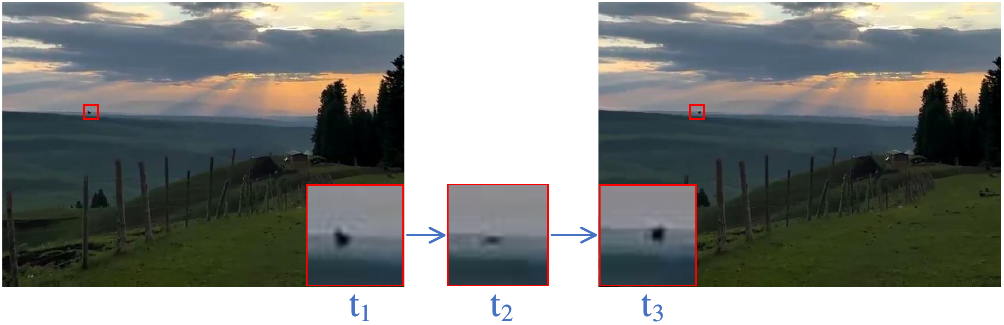}
	\caption{Challenges in motion detection of extremely tiny (ET-) targets: sparse visual cues, unstable outline, and background dynamics in a cluttered scenario. The figure illustrates a distance bird flying against a dynamic and cluttered background across time $t_{1}$, $t_{2}$, and $t_{3}$, with the bird highlighted and zoomed in in red boxes for better visibility. The bird occupies only few dozen pixels, like a spot, and its outline varies by flapping wings and flying attitude, making it difficult for motion detection.}
	\label{Fig:small_moving_target}
\end{figure}

\IEEEPARstart{M}{otion} detection for extremely tiny\footnotemark (ET-) targets is a fundamental low-level vision task, with its benefits to several downstream applications, including target tracking and motion segmentation, as well as real-world applications like automatic driving or anomaly detection. In essence, ET-target motion detection is a pixel-wise problem comprising two coupled sub-tasks: determining the location of motion (localization) and estimating its direction (orientation). Notably, the motion detection of such targets is inherently category-independent, as any distant object naturally diminishes to a few nondescript pixels, as demoed in Fig. \ref{Fig:small_moving_target}. Consequently, a robust system must distinguish minute moving entities from cluttered dynamic environments regardless of their original identity. Traditional feature-based approaches often fail under these conditions, as the extreme scarcity of visual cues and the presence of complex backgrounds undermine their effectiveness. Similarly, the lack of spatial information limits the performance of deep-learning pipelines, particularly in moving object proposals approaches \cite{kim2022learning, yasir2021review} for localization or optical flow methods \cite{teed2020RAFTRecurrent, wang2024SEARAFTSimple, sun2024StreamFlowStreamlined, dong2024MemFlowOptical, luo2024FlowDiffuserAdvancing, morimitsu2025DPFlowAdaptive} for orientation estimation.

\footnotetext{Following the common definition in pixels: large target $\in[96^{2}, \infty]$, medium target $\in[32^{2}, 96^{2}]$, small target $\in[16^{2}, 32^{2}]$, tiny target $\in[8^{2}, 16^{2}]$, and extremely tiny (ET-) target $\in[1, 8^{2}]$ \cite{ying2025VisibleThermalTiny}.}

Natural architectures with rich interpretability offer an alternative to heavy reliance on expert experience and knowledge of conventional and deep learning models \cite{zhai2021OpticalFlow, Fu2019review}. Recently, a remarkable example is the STMD architectures \cite{Wiederman2008ESTMD, Wang2020DSTMD, Wang2020STMDpuls, Wang2021time, Wang2022attention, ling2022mathematical, Xu2023frac, wang2024BioInspiredSmall, chen2024unveiling}. It is derived from the physiological Small Target Motion Detector neurons and corresponding neural pathways of insect visual systems, which dominate insects' intense responses to targets covering only $1^{\circ}-3^{\circ}$ of their visual field \cite{Nordstrom2006small, Nordstrom2006insect, Barnett2007retinotopic, Nordstrom2012neural}. However, the existing STMD architecture-based models only detect ET-target motion with a limited range of velocity, a shortcoming inherited by a delay-and-correlate mechanism \cite{HR1956EMD, BL1965EMD, Arenz2017temporal}. Furthermore, the existing directional STMD models, employing an isolated directional gradient calculation strategy, significantly compromise orienting accuracy and robustness across a broad range of velocities. As pointed out above, the velocity sensitivity of STMD models hinders their effectiveness in real-world applications, as targets in the physical world often exhibit varied velocity vectors, i.e., changing speed and direction.

To address the aforementioned velocity limitation, we propose vSTMD, a learning-free model featuring two primary innovations: cross-Inhibition Dynamic Potentials (cIDP) and Collaborative Directional Gradient Calculation (CDGC). The cIDP mechanism implements cross-inhibition between opposite-polarity channels, enabling the self-adaptive extraction of motion cues from ET-targets across a broad velocity spectrum. This process generates dual potentials that are subsequently correlated to locate ET-target motion from background noise. Complementing this, the CDGC represents the first collaborative directional gradient strategy that significantly enhances orientation estimation accuracy and robustness at varied velocities. Furthermore, the CDGC paradigm streamlines processing, reducing the computational overhead to one-eighth of that required by traditional directional STMD models.

Extensive experiments on the RIST dataset \cite{RIST_DataSet} demonstrate the superiority of the proposed framework: the core vSTMD exhibits robust localization and orientation capabilities on par with leading data-driven models. Moreover, the feedback-facilitated vSTMD-F variant further pushes the performance boundaries, achieving a $58\%$ relative improvement in the $mF_{1}$ score and a $39\%$ reduction in orientation error compared to state-of-the-art competitors. Remarkably, both models maintain exceptional efficiency exceeding 560 FPS, approximately 60 times faster than contemporary deep-learning models like DpFlow \cite{morimitsu2025DPFlowAdaptive}. The framework's generalization and practical utility are further validated through cross-task comparisons on the XS-VID dataset \cite{guo2024xs} and successful application in maritime search and rescue scenarios, underscoring its potential for high-precision, bio-inspired motion detection in complex, real-world environments.

To sum up, the contributions of this paper are fourfold:

\begin{itemize}
	\item \textbf{Self-adaptive Motion Integration}: We propose the cross-Inhibition Dynamic Potential (cIDP) mechanism. This self-adaptation component effectively captures salient motion cues for extremely tiny (ET-) targets across a broad velocity spectrum, addressing the velocity sensitivity limitations of traditional models.
	
	\item \textbf{Efficient Motion Directional Perception}: We present the first Collaborative Directional Gradient Calculation (CDGC), significantly enhancing orientation robustness and accuracy while reducing the computational overhead to merely one-eighth of previous architectures.
	
	\item \textbf{State-of-The-Art Performance}: The proposed vSTMD, powered by the cIDP and CDGC module, establishes a robust baseline for joint localization and orientation estimation on the RIST dataset. Its feedback-enhanced variant, vSTMD-F, further advances the state-of-the-art (SOTA) in the field, achieving a $58\%$ improvement in $mF_{1}$ score and a $39\%$ reduction in average angular error (AAE).
	
	\item \textbf{Superior Efficiency and Generalization}: This work revitalizes the STMD architecture by bridging the performance gap with deep learning while preserving its inherent speed advantage (achieving $560+$ FPS). Its broad generalization is further evidenced by superior performance in cross-task scenarios (XS-VID) and practical maritime search and rescue applications.
\end{itemize}

The rest of this paper is organized as follows. Sect. \ref{Sect:RelatedWork} reviews the related studies. Then, Sect. \ref{Sect:Preliminaries} attaches preliminaries, following the proposed method specified in Sect. \ref{Sect:Formular}. Next, Sect. \ref{Sect:Experiment} reports the experimental results, including effectiveness demonstration in systematic evaluations, comparisons, and ablations in the real-world dataset. Finally, some discussions and conclusions are given in the Sect. \ref{Sect:Discussion} and  \ref{Sect:Conclusion}, respectively.

\begin{figure*}[!t]
	\centering
	\includegraphics[width=1\textwidth]{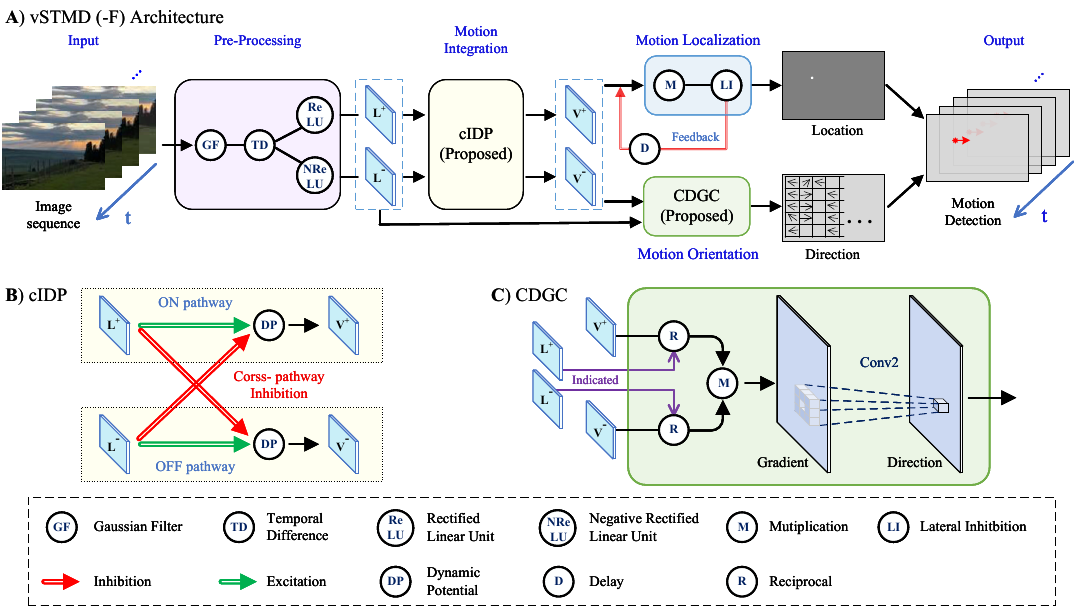}
	\caption{Schematic of the proposed work. (a) Overall architecture, comprising sequential stages of pre-processing, motion integration, localization, and orientation estimation. vSTMD refers to the baseline model, while vSTMD-F denotes the variant incorporating the feedback pathway \cite{Wang2021time}. (b) Detailed structure of the cross-Inhibition Dynamic Potential (cIDP) mechanism, where red double-line arrows indicate cross-inhibition pathways and DP represents the dynamic potential components. (c) The Collaborative Directional Gradient Calculation (CDGC) strategy employed for orientation estimation. In the schematic, circular elements represent pixel-wise operators, and warped cubes denote intermediate spatial feature maps that maintain the input image dimensions.}
	\label{Fig:Model_architecture}
\end{figure*}

\section{Related work}
\label{Sect:RelatedWork}

\subsection{Previous STMD Models}
Following the electrophysiological studies of hoverflies (see Sect. \ref{Sect:STMD_neuron}), Wiederman \emph{et al.} first proposed elementary STMD (ESTMD) to locate motion of ET-targets \cite{Wiederman2008ESTMD}. To further recognize motion direction, Wang \emph{et al.} (2020) proposed a directionally selective STMD (DSTMD) \cite{Wang2020DSTMD}. However, these two key backbones of the STMD architecture function for a single velocity as detailed in Sect. \ref{Sect:STMD_architecture}, due to the limitation inherited by the delay-and-correlate mechanism (see Sect. \ref{Sect:delay_and_correlate}). Thus, we propose novel mechanisms to address the single-velocity limitation of the previous STMD models. The main differences include: (1) for motion localization, we proposed a new cIDP for motion integration instead of the delay operator in ESTMD \cite{Wiederman2008ESTMD}; (2) for orientation, we present the first collaborative direction gradient calculation (CDGC) that replaces the isolated strategy (Isolated-DGC) in DSTMD \cite{Wang2020DSTMD}.

During the past decade, efforts have been made to improve the functionality and performance of STMD architectures. Notable advancements include a contrast pathway to remove contrast-invariable targets \cite{Wang2020STMDpuls}, a time-delay feedback pathway to capture fast-moving targets \cite{Wang2021time}, attention and prediction pathways to enhance multi-frame detection \cite{Wang2022attention}, a spatiotemporal feedback to suppress background motions \cite{wang2023bio}, and a Haar frequency domain to advance dim light performance\cite{chen2024unveiling}. In addition, some works mathematically support STMD architectures: studying feedback loop \cite{ling2022mathematical}, fractional-difference operator \cite{Xu2023frac}, and rigid propagation \cite{chen2025rigid}. In the proposed work, we adopt the fractional-difference operator by \cite{Xu2023frac} to extract temporal change in the pre-processing stage (Sect. \ref{Sect:preprocessing}) and employ the time-delay feedback pathway of \cite{Wang2021time} for the enhanced version vSTMD-F after the proposed cIDP. The remaining components are excluded from our work due to compatibility issues - their performance severely depends on the delay parameter.

\subsection{Optical Flow} \label{Sect:optic_flow}
Optical flow techniques, which estimate pixel-level displacement changes in low-level vision, appear effective for orientation estimation in motion detection tasks for ET-targets. However, traditional energy minimization approaches \cite{lucas1981IterativeImage} depend on consistent shape contours, making them unsuitable for ET-targets. Recent advances in deep learning \cite{teed2020RAFTRecurrent, wang2024SEARAFTSimple, sun2024StreamFlowStreamlined, dong2024MemFlowOptical, luo2024FlowDiffuserAdvancing, morimitsu2025DPFlowAdaptive} have significantly enhanced performance by leveraging superior feature extraction capabilities, achieving higher accuracy and faster processing speeds than conventional methods. Nevertheless, these feature-based approaches remain limited when visual cues are sparse or unstable, a frequent challenge in detecting ET-objects within cluttered and dynamic scenes. Furthermore, the scarcity of specialized datasets for ET-object motion hinders the training and optimization of optical flow methods \cite{alfarano2024EstimatingOptical}. 

\subsection{Moving Object Proposal}
Moving object proposal methods generate potential moving object candidates regardless of their inclusion in training data, comprising two classes: (1) associating objects between adjacent images via object proposal \cite{kim2022learning} and (2) segmentation of foreground and background \cite{yasir2021review}. For the first category, the effectiveness is significantly limited when processing ET-targets due to their minimal feature availability and unstable boundary characteristics across consecutive frames. The second category primarily relies on either optical flow estimation \cite{alfarano2024EstimatingOptical} (whose limitations for our specific task are detailed in Sect. \ref{Sect:optic_flow}) or background modeling techniques. However, background modeling approaches demonstrate poor performance in cluttered environments with dynamic scene elements.

\section{Preliminaries} \label{Sect:Preliminaries}
To better understand the idea and origin of modeling and provide supporting evidence for bio-plausibility, we present some preliminaries in this section.

\subsection{Small Target Motion Detector Neurons} \label{Sect:STMD_neuron}
Insects perform elaborate and elegant high-speed pursuits of prey, predators, or conspecifics, even when these objects are expressed as ET-targets in the insect's visual field. The tiny yet efficient brain of insects provides a tantalizing prototype for artificial modeling, attracting a lot of electrophysiological studies to reveal the principles behind these behaviors.

In hoverfly \textit{Eristalis tenax} \cite{Nordstrom2006small, Nordstrom2006insect, Barnett2007retinotopic} and dragonfly \textit{Hemicordulia} \cite{geurten2007NeuralMechanisms}, a specific class of neuron is recorded spiking strongly to moving targets within a visual angle of 1$^{\circ}$–3$^{\circ}$ while exhibiting minimal responsiveness to larger objects exceeding 10$^{\circ}$. Marvelously, these neurons can detect extremely tiny moving targets even without related motion cues, which are classified and named as small target motion detector neurons (STMDs). Similarly, two lobula column cells in the fruitfly \textit{Drosophila}, LC11 \cite{tanaka2020object} and LC18 \cite{klapoetke2022FunctionallyOrdered}, have been classified as pure small moving target sensitive neurons. In this paper, our model is mainly based on STMD architectures and refers to other pathways listed in Sect. \ref{Sect:neural_pathway}.

\subsection{Delay-and-Correlate Mechanism} \label{Sect:delay_and_correlate}

The delay-and-correlate mechanism, derived from the elementary motion detector (EMD) \cite{HR1956EMD}, is a foundational concept for understanding how physiological visual systems detect motion through compact and efficient networks, particularly in low-level pathways \cite{frye2015elementary}. Given an input $I(\vec{z}, t)$ and assuming $O_{model}(\vec{z}, t)$ as output, HR-EMD \cite{HR1956EMD} can be mathematically described as follows:
\begin{align}\label{Fomular:EMD}
	O_{HR}(\vec{z}, t) &= I(\vec{z'}, t-\tau) \cdot I(\vec{z}, t)
\end{align}
where $\vec{z}$ and $\vec{z'}$ represent different plane space coordinates, $t$ denotes time, and $\tau$ is a delay parameter. Even utilizing different forms, these methods all correlate the signals with and without delay, thus known as the delay-and-correlate mechanism.

However, despite its widespread use and historical significance, this mechanism only responds to a single velocity ($\hat{v} = |\vec{z}-\vec{z'}|\, / \, \tau$) of a signal for a preset $\tau$, based on the relationship between distance (correlated distance $|\vec{z}-\vec{z'}|$), speed (signal speed $\hat{v}$), and time ($\tau=t-(t-\tau)$). The existing STMD models incorporate a delay-and-correlate mechanism \cite{Wiederman2008ESTMD, Nordstrom2012neural}, thus inheriting its efficient yet shortcomings, as detailed in the next subsection, Sect. \ref{Sect:STMD_architecture}. In this paper, a novel dynamic-and-correlate methodology is proposed where cross-Inhibition Dynamic Potential (cIDP) acts as a self-adaptation mechanism in the motion integration stage (Sect. \ref{Sect:cIDP}). For correlation, we follow conventional multiplication for motion decriminalization and further design an indicated division for orienting recognition (see Sect. \ref{Sect:CDGC}).

\subsection{STMD Architecture} \label{Sect:STMD_architecture}

Current STMD architecture is derived from the delay-and-correlate mechanism, which detects motion for ET-targets by correlating two or more coincident input signals from ON (brightness increase) and OFF (brightness decrease) pathways \cite{Fu2019review, fu2023onoff}. The typical backbone of STMD models includes ESTMD \cite{Wiederman2008ESTMD} and DSTMD \cite{Wang2020DSTMD}, where only the latter is directionally sensitive. The later works \cite{Wang2020STMDpuls, Wang2021time, Wang2022attention, Xu2023frac, ling2022mathematical, wang2024BioInspiredSmall, chen2024unveiling} focus on improving performance in various scenarios based on these two backbones. Incidentally, two cascaded models, ESTMD-EMD and EMD-ESTMD \cite{Wiedermann2013biologically}, can also identify the motion direction, which is mathematically equivalent to DSTMD \cite{Wang2020DSTMD}.

For input stimulus $I(\vec{z}, t)$ with spatial coordinates $\vec{z}$ and time $t$, the STMD architecture mainly processes temporal changes and then divides them to ON (brightening) polarity channel $L^{+}(\vec{z}, t) = \left[L(\vec{z},t)\right]^{+}$ and OFF (darkening) polarity channel $L^{-}(\vec{z}, t) = \left[L(\vec{z},t)\right]^{-}$, based on polarity of temporal change $L(\vec{z},t) = \frac{d}{dt}I(\vec{z},t)$, where $[x]^{+}$ and $[x]^{-}$ are $\max(x, 0)$ and $\max(-x, 0)$, respectively. Next, the ESTMD \cite{Wiederman2008ESTMD} model correlates ON polarity and delayed OFF polarity to locate dark tiny targets, where the delay parameter $\tau$ is preset to match the target size $l$ along the motion path to the target velocity $v$. The DSTMD \cite{Wang2020DSTMD} extends ESTMD by correlating ON and OFF polarities in two spatial locations ($z$ and $z'$) with different delay parameters ($\tau, \tau'$, and $\tau''$) to locate dark tiny targets and simultaneously resolve their motion directions. Denoting outputs as $O_{\{model\}}$, it can be described as
\begin{subequations}\label{Fomular:STMD}
	\begin{align}
		O_{ESTMD}(\vec{z}, t) &= L^{+}(\vec{z},t) \cdot L^{-}(\vec{z},t-\tau) \\
		O_{DSTMD}(\vec{z}, t, \theta) &= L^{+}(\vec{z},t) \cdot L^{-}(\vec{z'},t-\tau) \nonumber\\
		& \cdot \left[L^{+}(\vec{z},t-\tau') 
		+ L^{-}(\vec{z'},t-\tau'') \right]  
	\end{align}
\end{subequations}
where $\vec{z'} = \vec{z} + \alpha \cdot (\cos \theta, \, \sin \theta)$.

As discussed above, neither ESMTD nor DSTMD backbones nor other STMD models necessitate the time parameter $\tau$ to make the inputs of ON and OFF polarities coincident in the output layer. As a result, the model only performs well when $\tau$ is close to $\frac{l}{v}$. In addition, DSTMD employs an isolated directional gradient calculation (Isolated-DGC) for orientation (Fig. \ref{Fig:advantage_direction}~(C-i)), which is time-consuming and susceptible to background motion. In real-world scenarios, however, it is challenging to establish an optimal parameter that can effectively accommodate the varying sizes and velocities of tiny targets. Therefore, we propose a new cIDP mechanism to collect motion cues and the first CDGC strategy for orientation, detailed in Sect. \ref{Sect:cIDP} and \ref{Sect:CDGC}. These two modules are both irrelevant to the delay parameter $\tau$ and thus better for real-world implementation, forming a methodology named dynamic-and-correlate.

\subsection{Insect Neural Pathway} \label{Sect:neural_pathway}
In the insect neural system, the sustained neuron Mi4 of the ON pathway and Tm9 of the OFF pathway can accumulate ipsilateral excitation, which respects motion edges of ET-targets with light and dark \cite{Arenz2017temporal}. Moreover, contralateral inhibition phenomena have been found in these two sustained neurons. Specifically, OFF neuron Tm9 receive inhibitory input from L1 (ON) via Mi4 \cite{takemura2013visual, fisher2015class}. Similarly, ON neuron Mi4 suggests an inhibitory effect from the OFF pathway, mediated by Mi9 \cite{Arenz2017temporal}. The above evidences support the proposed cIDP mechanism (Sect. \ref{Sect:cIDP}). 

In addition, the directional-insensitive pathway, L3-Tm9 \cite{fisher2015class}, inspires us to present a CDGC strategy (Sect. \ref{Sect:CDGC}). Moreover, \cite{klapoetke2022functionally} suggests that downstream neurons reading higher-order features are more likely to integrate inputs from neighboring upstream sources, supporting the direction gradient integration after CDGC for orienting. In addition, the indicated division in CDGC can be viewed as an inhibitory mechanism similar to the BL-detector \cite{BL1965EMD}, which is also utilized in the reversal of the preferred direction, as discussed in \cite{tanaka2022neural}.

\section{Methods and Formulations}\label{Sect:Formular}

The proposed vSTMD online locates and orients the motion of ET-targets from consecutive image sequences ${I(\vec{z}, t)}$, where $\vec{z}$ is the spatial index and $t$ denotes time. An overview of our approach is depicted in Fig. \ref{Fig:Model_architecture}. Our method can be distilled down to three stages: (1) pre-processing with spatial filter and temporal difference, (2) motion cues integration by the proposed cIDP, and (3) motion localization via correlation and orientation via the proposed CDGC. And last but not least, we illustrate the principle of localization and orientation for the proposed model, providing richer explainability compared to data-driven approaches, detailed in Sect. \ref{Sect:principle}.

\subsection{Pre-Processing} \label{Sect:preprocessing}

In the pre-processing stage, each input image of the sequence $I(\vec{z}, t)$ is first spatially blurred to denoise. This operator is counter-intuitive for image processing yet efficient in motion detection, which we demonstrate by ablation in the experiments section. Next, we adopt a fractional-difference operator $\mathscr{D}^{(\alpha)}_{t}$ \cite{Xu2023frac} to extract the temporal changes between adjacent blurred images, resulting in a change map ${L(\vec{z}, t)}$. Denoting blurred input as $P(\vec{z}, t)$, the above formulas are
\begin{align}
	P(\vec{z}, t) = I(\vec{z}, t) * G_{\sigma_{1}}(\vec{z}) 
	\label{Formula:Gauss_blur} \\
	L(\vec{z}, t) = \mathscr{D}^{(\alpha)}_{t} [ P(\vec{z}, t) ]
	\label{Formula:frac_diff}
\end{align}
where the sign $*$ denotes convolution, $G_{\sigma}(\vec{z})$ is a two-dimensional Gaussian function, and $\alpha$ represents the order of the fractional-difference operator.

Consequently, the change map ${L(\vec{z}, t)}$ are separated to ON and OFF polarities based on the sign of each pixel, denoting as $L^{+}(\vec{z}, t) = \left[L(\vec{z},t)\right]^{+}$ and OFF polarities $L^{-}(\vec{z}, t) = \left[L(\vec{z},t)\right]^{-}$, where $[x]^{+}$ and $[x]^{-}$ are $\max(x, 0)$ and $\max(-x, 0)$, respectively. Last, the two opposite polarities are fed into the next stage of motion integration.

\subsection{Motion Integration - cIDP} \label{Sect:cIDP}
To capture motion cues of ET-targets across a broad velocity range, we propose dynamics potentials incorporating ipsilateral excitatory and contralateral inhibition for the ON-OFF polarities, named cross-Inhibition Dynamics Potentials (cIDP). Specifically, the ipsilateral excitatory allows accumulation of the separated motion cues in ON-OFF pathways for light and dark motion edges across a multi-velocity range. To suppress motion cues of large objects and background clutter, contralateral inhibition is introduced between these opposing pathways, as shown in Fig. \ref{Fig:Motion_localization}~(C) and (D). The effective integration window is governed by the cross-inhibition term, enabling the neuron to aggregate evidence from both slow and fast stimuli.

The cIDP mechanism is calculated via the leakage-and-integrate membrane potential equation, which is well-established in computational neuroscience for modeling neuronal membrane potentials. Denoting ON and OFF potentials as $V^{+}$ and $V^{-}$, they can be formulated as:
\begin{subequations}\label{Fomular:cIDP}
	\begin{align}
		\frac{d}{dt} V^{+}(\vec{z},t) 
		= g_{L} \cdot \exp ^{L^{-}(\vec{z},t) } \cdot [v_{R} - V^{+}(\vec{z},t)] 
		\nonumber \\
		+ L^{+}(\vec{z},t) \cdot [v_{E} - V^{+}(\vec{z},t)]
		\label{Formula:ON_current}
		\\
		\frac{d}{dt} V^{-}(\vec{z},t) 
		= g_{L} \cdot \exp^{L^{+}(\vec{z},t)} \cdot [v_{R} - V^{-}(\vec{z},t)] 
		\nonumber \\
		+ L^{-}(\vec{z},t) \cdot [v_{E} - V^{-}(\vec{z},t)]
		\label{Formula:OFF_current}
	\end{align}
\end{subequations}
where $g_{L}$, $v_{R}$, and $v_{E}$ are the leakage coefficient, resting potential, and excitatory reversal potential, respectively. Notably, here we employ a variant of a shunting inhibition-type \cite{hodgkin1952quantitative} as the graded potential \cite{haag1997encoding, simmons1999performance} characteristics, and hyperpolarization is excluded due to non-spike \cite{zettler1973active, hengstenberg1977spike}, observed in the insects' low-level system.

\subsection{Motion Localization and Orientation} \label{SubS:direction}

In this part, we separate localization and orientation, rather than the one-stage calculation of localization and orientation in the previous directional STMD. 

\subsubsection{Motion Localization}

For motion localization, we follow second-order correlation in \cite{Wiederman2008ESTMD}, where the dynamics potentials of $V^{+}$ and $V^{-}$ are multiplied to pixel-wise determine if a motion is caused by an ET-target. vSTMD output confidence coefficient ($O_{cc}(\vec{z}, t)$) of each pixel, after a lateral inhibition for suppressing tiny bars. As shown in Fig. \ref{Fig:Model_architecture}~(B-i), $O_{cc}(\vec{z}, t)$ can be described as follows,
\begin{align}
	O_{cc}(\vec{z}, t) = [V^{+}(\vec{z}, t) \cdot V^{-}(\vec{z}, t)] * W(\vec{z}).
	\label{Formula:discrimination}
\end{align}
where the sign $*$ denotes the convolution operator. $W(\vec{z})$ is a two-dimensional Mexican hat kernel, denoted as
\begin{subequations}
	\begin{align}
		W(\vec{z}) = A \left[ g(\vec{z})\right]^{+} - B \left[g(\vec{z})\right]^{-}
		\label{Fomular:AB} \\
		g(\vec{z}) = G_{\sigma_{2}}(\vec{z}) - e \cdot G_{\sigma_{3}}(\vec{z}) -\rho
		\label{Fomular:erho}
	\end{align}
\end{subequations}
where $G_{\sigma}(\vec{z})$ is a two-dimensional Gaussian kernel, and $A$, $B$, $e$, $\rho$, $\sigma_{1}$, and $\sigma_{2}$ are constants.

Building upon the correlation mechanism in the baseline vSTMD (Formula \ref{Formula:discrimination}), we incorporate a feedback pathway \cite{Wang2021time} in the improved variant vSTMD-F, to extend the velocity range of localization (verified in Fig. \ref{Fig:v2AUC}~(D)), while preserving the original orientation estimation framework. Let $\tilde{O}_{cc}(\vec{z},t)$ denote the motion localization confidence coefficient for vSTMD-F, which can be described as
\begin{align}
	&\tilde{O}_{cc}(\vec{z},t)  \nonumber \\
	=& \{[V^{+}(\vec{z},t) - F(\vec{z},t)] \cdot  [V^{-}(\vec{z},t) - F(\vec{z},t)]\} * W(\vec{z}).
	\label{Formula:discrimination_feedback}
\end{align}
where $W(\vec{z})$ is determined by Formula (\ref{Fomular:AB}) and (\ref{Fomular:erho}). The feedback signal $F(\vec{z},t)$ is calculated using a two-dimensional Gaussian kernel $G_{\sigma}(\vec{z})$:
\begin{align}
	F(\vec{z},t)  = \beta \cdot [ \tilde{O}_{cc}(\vec{z},t-\Delta t) + O_{cc}(\vec{z},t-\Delta t) * G_{\sigma_{4}}(\vec{z}) ]
	\label{Formula:feedback_pathway}
\end{align}
where $\beta$ represents the feedback contrast. 

Notably, we implement an instantaneous feedback mechanism (operating from time $t-\Delta t$ to $t$) rather than relying on the specific temporal delay parameters typically required by traditional feedback pathways \cite{Wang2021time}. The feasibility of this undelayed feedback approach is further validated by the ablation studies presented in Sect. \ref{Sect_Ablation}.

\subsubsection{Motion Orientation - CDGC} \label{Sect:CDGC}

We propose Collaborative Directional Gradient Calculation (CDGC), the first strategy that perceives collaborative gradient for motion orientation of ET-targets, tailored to the cIDP. The CDGC brings two benefits: (1) Collaborative gradient greatly reduces computational load, with only one eighth of the previous directional models. (2) CDGC enhances accuracy and robustness by focusing on motion cues within targets. 

Denoting the direction gradient as $G(\vec{z}, t)$, it can be described via a piece-wise function:
\begin{equation}
	G(\vec{z}, t) = 
	\begin{cases}
		V^{-}(\vec{z}, t) / V^{+}(\vec{z}, t) ,
		& \text{if} \ L^{+}(\vec{z}, t) > 0 \\
		V^{+}(\vec{z}, t) / V^{-}(\vec{z}, t) ,
		& \text{if} \ L^{-}(\vec{z}, t) > 0 \\
		0 ,
		& \text{otherwise}.
	\end{cases}
	\label{Formula:division}
\end{equation}
Here, the equation (\ref{Formula:division}) can be regarded as an identified division correlation between $V^{+}(\vec{z}, t)$ and $V^{-}(\vec{z}, t)$ with indicator of $L^{+}(\vec{z}, t)$ (i.e., $L(\vec{z}, t) > 0$) and $L^{-}(\vec{z}, t)$ (i.e., $L(\vec{z}, t) < 0$), which is equivalent to multiply the numerator by the reciprocal of the denominator, as shown in Fig. \ref{Fig:Model_architecture}~(C) and Fig. \ref{Fig:motion_orientating}~(F-G). In this way, the motion direction $O_{d}(\vec{z}, t)$ of ET-targets is encoded by a transition from low to high gradients and thus can be decoded by fusing the surrounding gradients:  
\begin{align}
	O_{d}(\vec{z},t)
	= \arctan \left(
	\frac{\int_{0}^{2\pi} G(\vec{z}+(\cos\theta, \sin\theta),t) \cdot \sin\theta \ d\theta}
	{\int_{0}^{2\pi} G(\vec{z}+(\cos\theta, \sin\theta),t) \cdot \cos\theta \ d\theta}
	\right)
	\label{Formula:direction}
\end{align}
Notably, the CDGC is tailored for ET-targets as the orientation $O_{d}(\vec{z},t)$ is only correct within the area marked by motion localization $O_{cc}(\vec{z},t)$. Specifically, for small targets, the leading and trailing edges are temporally close, and the corresponding ON–OFF channels produce well-aligned, high-contrast signals. This alignment ensures that the motion gradient is correctly captured, and the orientation estimation within the localized target area is reliable. In contrast, for large targets, the leading edge and trailing edge are spatially separated by a larger distance. By the time the trailing edge enters the same spatial location, the activity from the leading edge has already decayed. Consequently, the ON–OFF channels no longer encode the true motion gradient of the target, which is instead dominated by other background or residual signals. As a result, the orientation becomes unreliable outside the localized small-target region.

\subsubsection{Final Output}
In summary, vSTMD-F generates pixel-wise outputs $[\tilde{O}_{cc}, O_{d}]$ (computed via Formula (\ref{Formula:discrimination_feedback}) and (\ref{Formula:direction})) representing the confidence coefficient for motion localization of ET-targets and their corresponding orientation. In comparison, the baseline vSTMD produces outputs $[O_{cc}, O_{d}]$ through Formula (\ref{Formula:discrimination}) and (\ref{Formula:direction}).

\begin{figure}[!t]
	\centering 
	\includegraphics[width=0.5\textwidth]{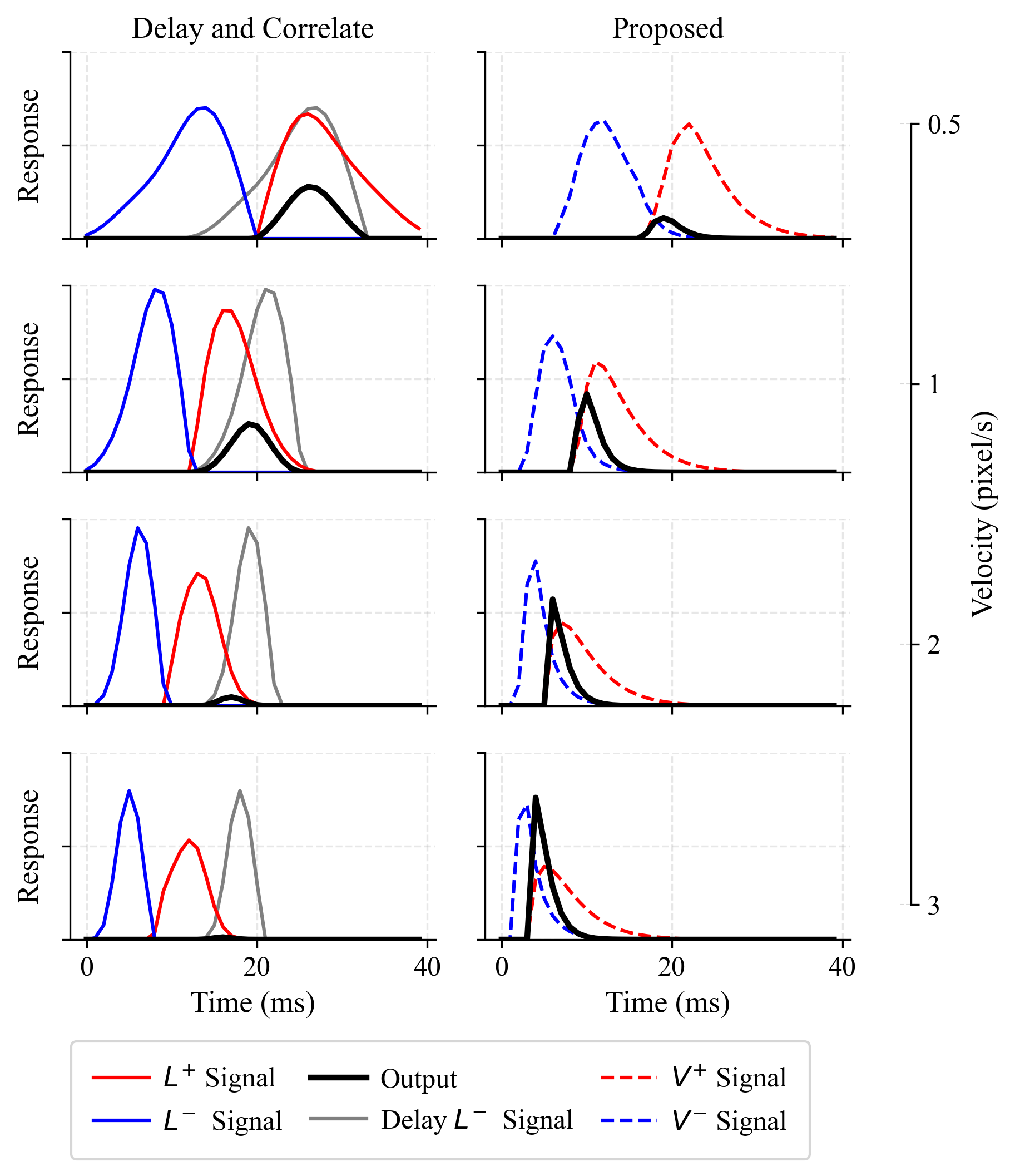}
	\caption{Output comparison between the conventional delay-and-correlate mechanism (left column) and the proposed model (right column) at velocities ranging from 0.5 pixel per frame in the top row to 3 pixel per frame in the bottom row. The $L^{+}$ and $L^{-}$ signals denote the ON and OFF motion edges produced by extremely tiny targets. In the left column, the delay-and-correlate mechanism imposes a fixed temporal delay between the ON and OFF channels to ensure temporal correct correlation. As velocity increases, the delayed OFF signal becomes increasingly misaligned with the ON signal. Such misregistration shifts the response away from the target - outside the summits of ON and OFF motion edges, and reduces the response's amplitude. In contrast, the right column illustrates the proposed model’s self-adaptation across various velocities. The dynamic potentials of the proposed cIDP, $V^{+}$ and $V^{-}$, calculated by Formula (\ref{Formula:ON_current}) and (\ref{Formula:OFF_current}), steadily exhibit opposite gradients within targets, and their product yields robust responses across velocities and remains consistently localized within the actual target.}
	\label{Fig:various_velocity_localization}
\end{figure}

\subsection{Superiority and Interpretability}\label{Sect:principle}

Here, we demonstrate the superiority with schematic diagrams concerning the various velocity localization as well as robust and efficient motion orientation. Moreover, we visualize the processing pipeline to demonstrate our model’s principles through one-dimensional signal transformations from input to output. These demonstrations offer greater explainability than deep learning-driven methods, as the proposed model aligns more naturally with interpretable architectures.

\subsubsection{Board-Velocity Range Motion Localization}

Here, we compare the conventional delay-and-correlate mechanism with our model over a wide range of target velocities. Figure \ref{Fig:various_velocity_localization} shows the responses of both methods when the target moves from 0.5 to 3 pixels per frame, where the target size is $5 \times 5$ pixels. For extremely tiny targets, the ON and OFF pathways generate a pair of motion-edge responses, denoted as $L^{+}$ and $L^{-}$, where a more detailed pipeline and explanation for these variables are shown in Fig. \ref{Fig:Motion_localization}.

In the delay-and-correlate mechanism, a fixed temporal offset is imposed between the ON and OFF signals. As the velocity increases, this fixed delay causes the delayed OFF signal to drift progressively away from the true ON–OFF edge alignment. Consequently, the correlation output becomes shifted outside the target region, and its magnitude deteriorates, making the response increasingly unreliable at high velocities.

In contrast, the proposed model introduces a cross-inhibition–driven dynamic potential (cIDP), which allows the ON and OFF channels to adjust their relative temporal profiles automatically. The resulting potentials, $V^{+}$ and $V^{-}$, develop opposite gradients around the target. Their product, therefore, yields a stable and well-localized response that remains confined within the target across all tested velocities. This dynamic and adaptive mechanism enables the model to maintain reliable detection because fast motions generate strong instantaneous potentials, whereas slow motions accumulate over time to reach comparable levels. Therefore, the model naturally maintains strong responses across a wide range of target velocities, without adjusting any timing parameters.

\subsubsection{Temporal Schematic for Motion Localization} 
\begin{figure}[!t]
	\centering
	\includegraphics[width=0.5\textwidth]{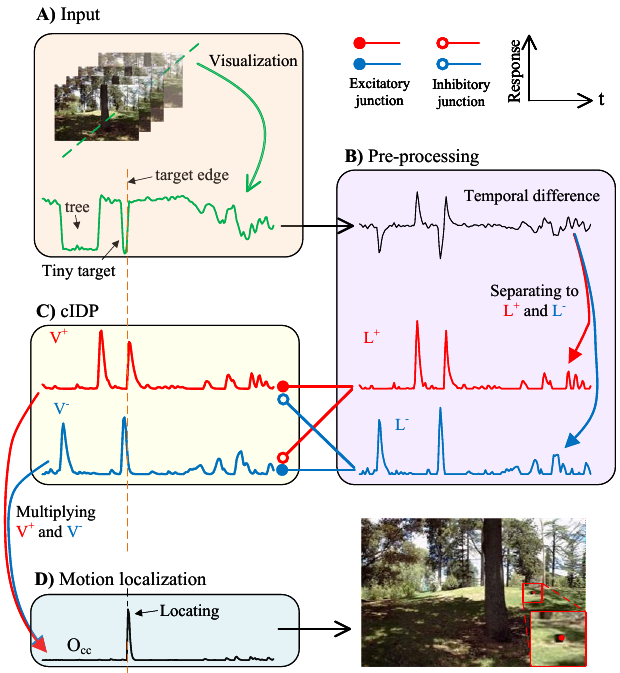}
	\caption{The temporal signal processing schematic for motion localization. Panel (A) showcases a signal composed of a tiny target against a cluttered and dynamic background, recorded for a fixed pixel along the temporal axis from a set of images. In (B), temporal changes of the input signal are extracted and then separated into $L^{+}$ and $L^{-}$ based on brightness increases and decreases. Next, panel (C) illustrates the output of the cross-Inhibition Dynamic Potentials (cIDP) mechanism of ON ($V^{+}$) and OFF ($V^{-}$) with ipsilateral excitatory and contralateral inhibition junctions. Finally, these two potentials are multiplied to locate the motion of ET-targets in (D), while there is no response to the dynamic background or larger objects such as the tree.}
	\label{Fig:Motion_localization}
\end{figure}

Fig. \ref{Fig:Motion_localization} demonstrates the motion localization mechanism of our proposed vSTMD for ET-targets\footnote{For comparison, the principle of conventional delay-and-correlate-type motion localization can be found in \cite{Wiederman2008ESTMD}.}. In (A), we record a temporal signal in a fixed pixel form image sequence, including an ET-target moving independently against a cluttered and dynamic background. Next, the input is decomposed into $L^{+}$ and $L^{-}$ during the pre-processing stage after the temporal difference. Then, panel (C) showcases the results of Formula (\ref{Formula:ON_current}) and (\ref{Formula:OFF_current}). In this procedure, ipsilateral excitation independently accumulates motion cues within ON-OFF pathways, while contralateral inhibition suppresses interference from large objects (for example, the tree in panel(A)) and background dynamics before multiplying correlation. Finally, the dual dynamic potentials $V^{+}$ and $V^{-}$ are multiplied to achieve motion selectivity for ET-targets. As shown in panel (D), a strong response occurs exclusively to the ET-target. Unlike the delay-and-correlate STMD, our approach does not require a preset parameter for velocity and works well across velocities.

\subsubsection{Robust, Accurate, and Efficient Motion Orientation} \label{Sect:orientation_pipeline}
\begin{figure}[!t]
	\centering 
	\includegraphics[width=0.5\textwidth]{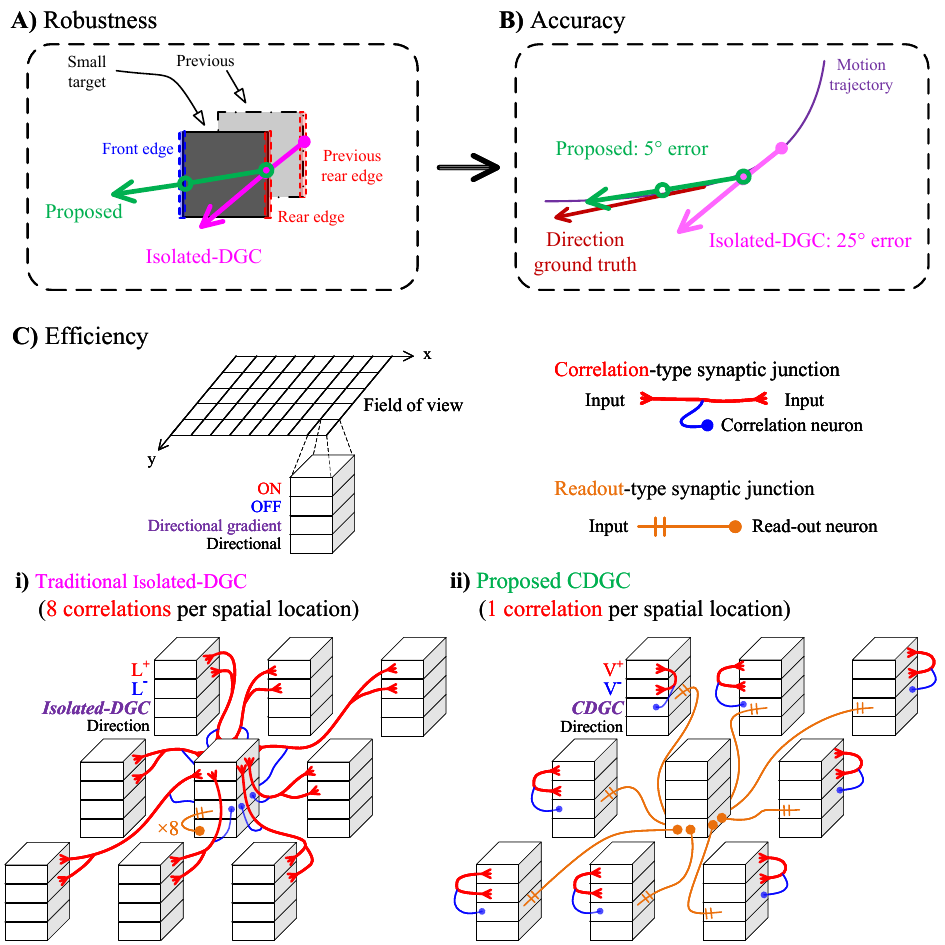}
	\caption{Robustness, accuracy, and efficiency of the proposed CDGC compared with traditional Isolated-DGC \cite{Wang2020DSTMD}. (A) illustrates the robustness of the proposed CDGC (green arrows) by only leveraging front and rear motion edges (green hollow circle). In comparison, the Isolated-DGC (pink arrows) calculates correlations between a current motion edge and a previous motion edge outside the current target (solid pink circle), making it susceptible to interference from background motion. (B) shows the high directional accuracy of the proposed CDGC. (C) highlights the efficiency of the proposed CDGC over the conventional Isolated-DGC. For simplicity, we illustrate a central location and eight surrounding locations, along with how directional information can be calculated, where the neuron names are indicated in the upper left corner. In the proposed CDGC, direction information can be directly read out from the sharing gradient without additional computation. Consequently, the proposed CDGC requires only one correlation per spatial location, while the traditional Isolated-DGC necessitates eight correlations between ON and OFF signals at each spatial location.}
	\label{Fig:advantage_direction}
\end{figure}
The advantages of the proposed CDGC strategy are demonstrated in Fig.~\ref{Fig:advantage_direction}, highlighting its robustness, accuracy, and efficiency. Here we compare CDGC to the isolated direction gradient calculation (Isolated-DGC) \cite{Wang2020DSTMD}, which is a variant of the delay-and-correlate mechanism detailed in Sect. \ref{Sect:delay_and_correlate}. As shown in panels (A) and (B), the CDGC exhibits robustness and accuracy by relying solely on information from the current target, specifically the front and rear motion edges (green hollow circles). In contrast, the Isolated-DGC (indicated by the pink arrows) calculates correlations between a current motion edge and a previous one outside the current target (represented by solid pink circles). This method exhibits poor robustness and huge orientation errors because searching for the prior motion edge in multiple directions introduces background interference. In addition, (C) highlights the computational efficiency of the two strategies by comparing their correlation requirements. Sub-panel (i) shows the traditional Isolated-DGC, which requires \textbf{eight correlations} per spatial location, resulting in a dense network of connections between "ON" and "OFF" channels. In contrast, the proposed model needs \textbf{only one correlation} per spatial location, significantly simplifying the connection complexity, as shown in sub-panel (ii).

\subsubsection{Spatial Schematic for Motion Orientation}
\begin{figure}[!t]
	\centering
	\includegraphics[width=0.5\textwidth]{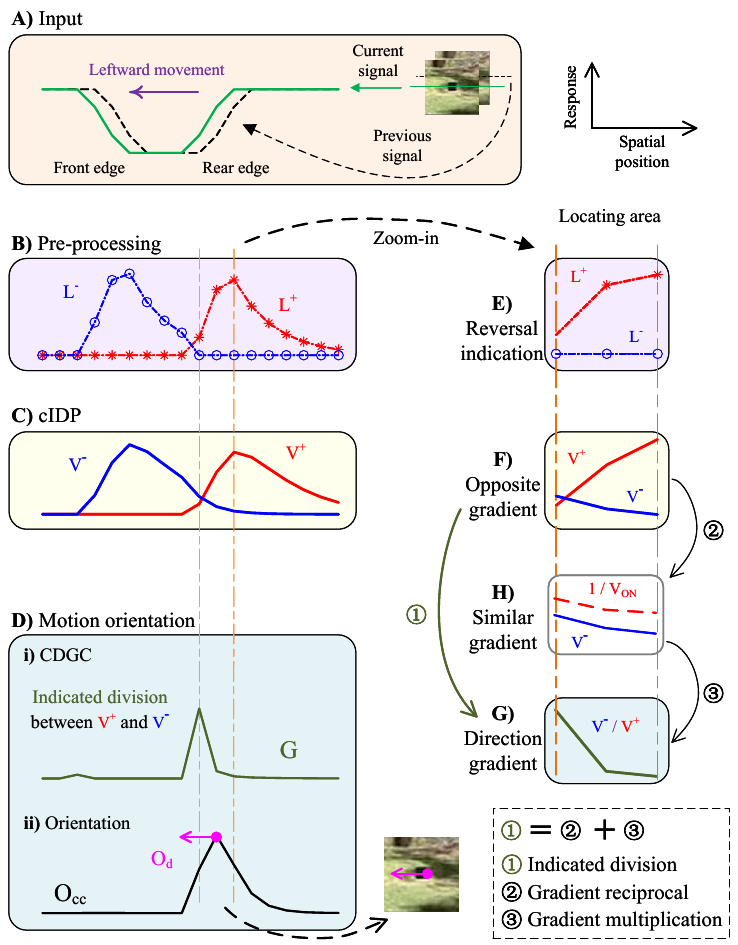}
	\caption{Schematic diagram of CDGC for motion orientation. In panel (A), we record a line of the current frame of a leftward-moving dark tiny target (green solid line), with the previous frame input shown by a black dotted line. (B) and (C) showcase the pre-processing and cIDP. In panel (D), CDGC calculates directional gradient $G$ (denoted by a blackish green solid line), where the target moves from a low gradient to a high gradient \textbf{within the locating area} (both $V^{+}$ and $V^{-}$ are nonzero, between two orange dotted lines). Finally, orientation $O_{d}$ can be recognized by fusing the surrounding gradient, indicated by the purple arrow, where the black solid line represents motion localization $O_{cc}$ in (ii). Furthermore, to better explain, we zoom in on the (E)-(G) and add an equivalent sub-process in (H) to represent the idea of indicated division, making the gradient consistent via reciprocal. Specifically, the indicated division reverses the spatial gradient of the rear edge in accordance with that of the front edge (H), where the indication is from (E). Finally, the directional gradient in (G) reflects the motion trend.}
	\label{Fig:motion_orientating}
\end{figure}

For the ET-target, motion direction is conveyed from the rear edge to the front edge and is implied within the ON and OFF edges. To orient a motion, we first calculate the directional gradient. To better illustrate, we recorded a spatial signal with a leftward movement of an ET-target and processed it step-by-step.

In Fig. \ref{Fig:motion_orientating}, panels (A - C) depict the dual-dynamics calculation with the same process in Fig. \ref{Fig:Motion_localization}~(A-C), but notably Fig. \ref{Fig:motion_orientating} illustrate spatial index rather than temporal axis, while panels (E-G) zoom in on the locating area. From panel (F), we observe that the spatial gradients of dual dynamics are opposite between the motion of front and rear edges within the locating area of a tiny moving target (marked between the two orange dotted lines where $V^{+}$ and $V^{-}$ are both positive). Specifically, along the motion direction, the gradient of the front edge increases, while that of the rear edge decreases, irrespective of their polarity. By reversing the spatial gradient of the rear edge to align it with the front edge in panel (H), directional gradients are created by combining the reversed rear edge and the original front edge in (G).

The CDGC mechanism utilizes $L^{+}$ and $L^{-}$ to identify the polarity of the target's rear edge. Specifically, within the localized regions, $L^{+} > 0$ indicates that the ON edge serves as the rear edge, necessitating a reversal of $V^{+}$. Otherwise, $V^{-}$ is reversed. By using $L^{+}$ and $L^{-}$ as polarity indicators, CDGC aligns the spatial gradient of the rear edge with that of the front edge (H), ensuring consistent directional coding for both light and dark targets. Notably, a reciprocal operator is employed as the reversal mechanism instead of a negation to ensure that the directional gradient remains non-negative.

\section{Experimental Results} \label{Sect:Experiment}
In this section, we first outline the experimental setup, followed by effectiveness verification. Then, we present comparisons and ablation. Next, the module time cost analysis is showcased. Additionally, the parameter analysis and response curves are listed in the Sect. I-C and I-D of \textbf{Supplementary Materials}.

\begingroup
\renewcommand{\arraystretch}{1.2} 
\begin{table}
	\centering
	\caption{Parameters of the proposed work.}
	\begin{tabular}{c c} 
		\toprule 
		Eq. & Parameters \\
		\midrule
		(\ref{Formula:Gauss_blur})	&
		$\sigma_{1} = 1$  \\
		(\ref{Formula:frac_diff})	&
		$\alpha = 0.25$   \\
		(\ref{Formula:ON_current}), (\ref{Formula:OFF_current})	&
		$g_{l} = 0.35, V_{r} = 0, V_{e} = 1$	\\
		(\ref{Fomular:AB})	&
		$A = 1, B = 3$	\\
		(\ref{Fomular:erho})	&
		$\sigma_{2} = 5, \sigma_{3} = 10, e = 3, \rho = 0$	\\
		(\ref{Formula:feedback_pathway})	&
		$\beta = 1, \sigma_{4} = 1.5$	\\
		\bottomrule
	\end{tabular}
	\label{Tab:Parameters}
\end{table}
\endgroup

\subsection{Experimental Setup}

\subsubsection{Datasets}
This study employs two distinct datasets for evaluation: simulated datasets generated using VisionEgg~\cite{Straw2008vision} and real-world datasets referred to as RIST~\cite{RIST_DataSet}. Both datasets feature dynamic backgrounds influenced by ego-motion, providing a challenging environment for model evaluation.

The synthetic datasets consist of multiple videos with a resolution of $470 \times 310$ pixels. Each video depicts a synthetic target moving against a built-in panoramic background. These synthetic targets vary in luminance, size, and motion speed and direction, enabling a comprehensive assessment of the proposed model's performance under diverse conditions.

The RIST dataset \cite{RIST_DataSet} comprises 19 videos totaling 29,900 frames, each with a resolution of $480 \times 270$ pixels and a frame rate of 240 Hz. The videos feature moving targets with variable-speed motion against diverse (listed in Sect. I-A of \textbf{Supplementary Materials}), cluttered backgrounds such as skies, buildings, trees, grass, and shrubs, under both sunny and cloudy conditions. This variety provides a rich foundation for evaluating the model's robustness and generalizability in real-world scenarios. Some representative frames are shown in the first row of Fig. \ref{Fig:result_in_RIST}.

\subsubsection{Comparative Models}
The proposed method is evaluated against a diverse set of competitive baselines, ranging from classical STMD models to advanced biologically inspired architectures based on Drosophila. Furthermore, to provide a rigorous assessment of directional estimation, we include several SOTA deep-learning-based optical flow models. These models serve as performance benchmarks for motion orientation, representing the cutting-edge capabilities of current motion analysis frameworks.

\begin{itemize}
	\item Classical STMD models. We select seven biologically inspired STMD variants, including ESTMD \cite{Wiederman2008ESTMD}, DSTMD \cite{Wang2020DSTMD}, Frac-STMD \cite{Xu2023frac}, STMD+ \cite{Wang2020STMDpuls}, ApgSTMD \cite{Wang2022attention}, FeedbackSTMD \cite{Wang2021time}, and $\mathcal{F}$STMD \cite{ling2022mathematical}. Among them, DSTMD, STMD+, and ApgSTMD can estimate both localization and orientation, while the remaining models produce localization only. These methods represent the mainstream delay-and-correlate paradigm and serve as the standard baselines for STMD architecture-based motion detection.

	\item Natural architectures in \textit{Drosophila}. To further compare with biologically grounded motion-sensitive neurons beyond STMD pathways, we also include small-object displacement detectors LC11 \cite{tanaka2020object} and LC18 \cite{klapoetke2022FunctionallyOrdered}, whose neuronal circuits originate from the fruit fly visual system. 
	
	\item Optical-flow-based models. To establish a robust benchmark for directional accuracy, we incorporate six SOTA deep-learning-based flow estimators: RAFT \cite{teed2020RAFTRecurrent}, SEA-RAFT \cite{wang2024SEARAFTSimple}, StreamFlow \cite{sun2024StreamFlowStreamlined}, MemFlow \cite{dong2024MemFlowOptical}, FlowDiffuser \cite{luo2024FlowDiffuserAdvancing}, and DPFlow \cite{morimitsu2025DPFlowAdaptive}. These models prioritize global motion fields over small-target selectivity, thus providing only orientation-specific baselines.

\end{itemize}

\subsubsection{Evaluation Criteria}
In our evaluation, we assess detection quality using the AUC (Area Under the ROC Curve), AR (Average Recall), AP (Average Precision), and F$_{1}$-score for localization, while the AAE (Average Angular Error) is employed to evaluate orientation. Additionally, we measure computational efficiency using the FPS (frames per second) on both CPU and GPU.

\subsubsection{Implementation Details}

The parameter settings for the proposed model are listed in Table \ref{Tab:Parameters}. All compared methods are implemented based on their official codes or with the recommended configurations. To avoid redundancy during the evaluation process, Non-Maximum Suppression (NMS) is applied to the outputs of all models.

All experiments presented in this paper were conducted using Python 3.12 and MATLAB R2024b on a personal laptop equipped with a 2.50 GHz Intel Core i5-12500 CPU with 32 GB of DDR5 memory, and an NVIDIA GeForce RTX 3050 Laptop GPU with 4GB of GPU memory. The implementation of the STMD models was carried out using the open-source toolbox Small-Target-Motion-Detectors (STMD version 2.0)~\cite{STMDgit}. The source code is available at \url{https://github.com/MingshuoXu/vSTMD}.

\subsection{Various Velocities Motion Detection Effectiveness} 

To systematically evaluate the effectiveness of our model across a wide dynamic range of motion, we benchmark its performance on a synthetic dataset with target velocities ranging from $0.1$ to $10$ pixels per frame (pixels/frame). This carefully selected range allows us to assess detection capabilities across two critical motion regimes: sub-pixel displacements, which require high sensitivity to subtle changes, and high-speed motion exceeding the target’s own spatial dimensions ($5 \times 5$ pixels), which poses challenges related to temporal aliasing and motion blur.

\subsubsection{Localization}
\begin{figure}[!t]
	\centering
	\includegraphics[width=0.48\textwidth]{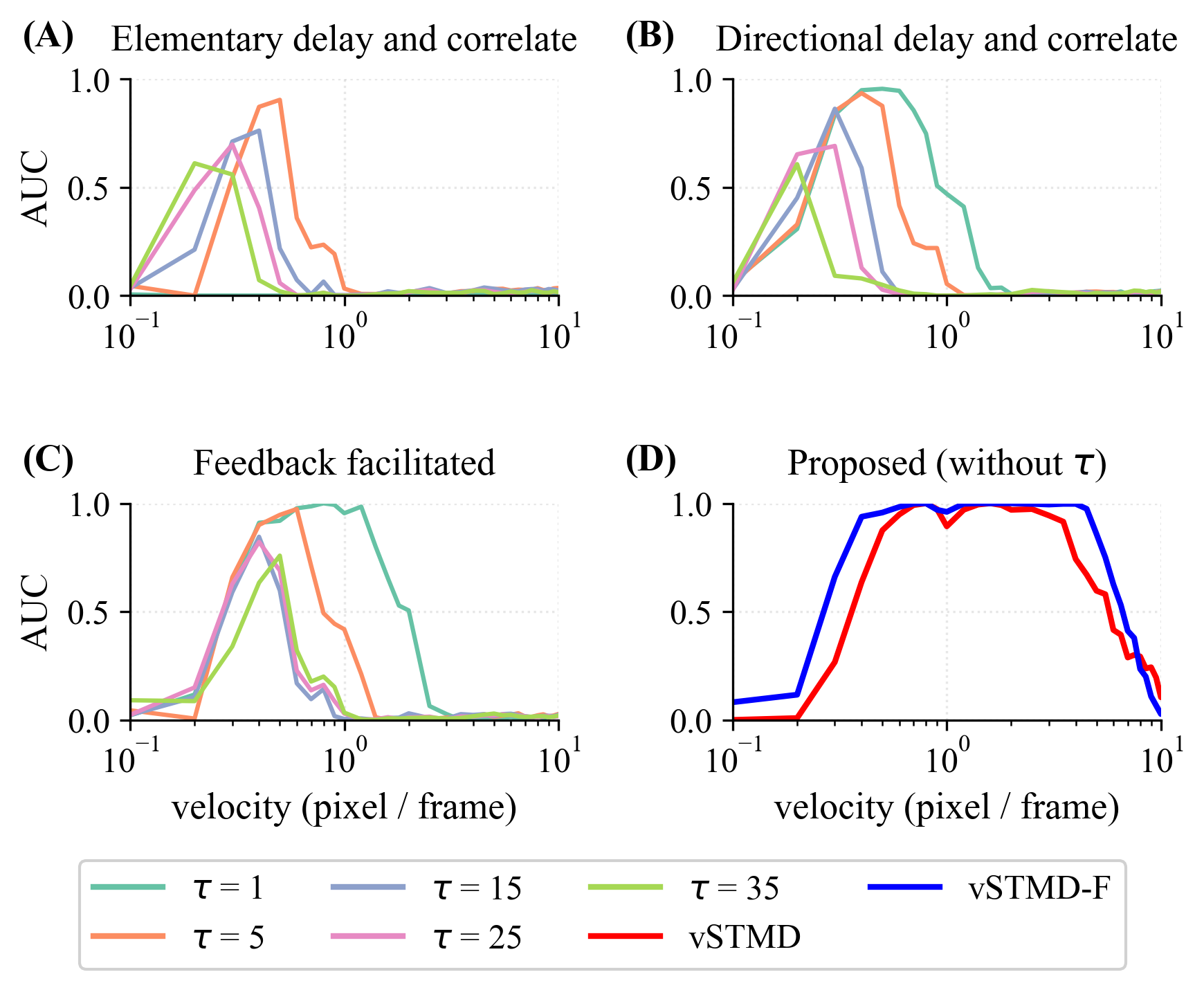}
	\caption{Velocity-AUC performance curves across varying motion velocities. (A)–(C) Three types of STMD models, representing elementary \cite{Wiederman2008ESTMD}, directional \cite{Wang2020DSTMD}, and feedback-facilitated \cite{Wang2021time} delay-and-correlate mechanisms, respectively. These models exhibit narrow effective operating ranges that are strictly dependent on the delay parameter $\tau$. (D) Performance of the proposed vSTMD and its feedback-enhanced variant (vSTMD-F), demonstrating robust, velocity-invariant detection across an extensive velocity spectrum without requiring $\tau$ adjustment.}
	\label{Fig:v2AUC}
\end{figure}

To evaluate velocity sensitivity in localization, we benchmark the proposed mechanism against three STMD models: ESTMD \cite{Wiederman2008ESTMD}, DSTMD \cite{Wang2020DSTMD}, and FeedbackSTMD \cite{ling2022mathematical} representing elementary, directional, and feedback facilitated delay-and-correlate mechanisms, respectively. We focus on how the delay parameter ($\tau$) dictates the velocity tuning of the delay-and-correlate mechanism. 

As shown in Fig. \ref{Fig:v2AUC}~(A–C), the delay-and-correlate mechanisms exhibit a rigid dependency on $\tau$: (1) their peak velocity responses shift toward higher velocities, and (2) their velocity tuning bandwidth widens. These results confirm that a fixed delay-and-correlate unit acts as a velocity-tuned filter, necessitating manual parameter setting for different motion scales. 

In contrast, Fig. \ref{Fig:v2AUC}~(D) demonstrates that the proposed vSTMD sustains high AUC performance across a wide dynamic range of velocities without $\tau$ adjustment. The integration of feedback further expands this effective operating range. These findings verify that the proposed cIDP module serves as a robust, self-adaptive mechanism capable of maintaining detection efficacy across diverse velocity profiles.

\subsubsection{Orientation}

\begingroup
\renewcommand{\arraystretch}{1.2} 
\begin{table}[t]
	\centering
	\caption{Performance comparison on the panoramic dataset in terms of Absolute Angular Error (AAE, rad $\downarrow$). The target size is $5 \times 5$ pixels. Velocity ($v$) is measured in pixels per frame. Directions with AAE $> 1.57$ ($\pi/2$) are considered unreliable and marked in {\color{gray}gray}. The average rank is reported in the rightmost column. The best and second-best results for each velocity are highlighted in \textbf{bold} and \underline{underlined}, respectively.}
	\begin{tabular}{r|ccccc|c} 
		\toprule
		\diagbox{Method}{AAE$\downarrow$}{Velocity} 
		& \parbox{0.6cm}{\centering 0.1\\$\shortmid$\\0.5} 
		& \parbox{0.6cm}{\centering 0.6\\$\shortmid$\\1} 
		& \parbox{0.6cm}{\centering 1.1\\$\shortmid$\\3}
		& \parbox{0.6cm}{\centering 3.1\\$\shortmid$\\5} 
		& \parbox{0.6cm}{\centering 5.1\\$\shortmid$\\10}
		& \parbox{0.6cm}{\centering Avg. Rank} \\
		\midrule
		DSTMD\cite{Wang2020DSTMD} & 
		\textbf{0.65}  & 0.53  & - & - & - & 9 \\
		STMD+\cite{Wang2020STMDpuls} & 
		\underline{0.69}  &   0.67   &  1.44  &  {\color{gray}1.75} &  {\color{gray}1.73} & 10\\
		ApgSTMD\cite{Wang2022attention} & 
		0.75 &   0.71 &  1.17 &  1.57 & {\color{gray}1.59} & 7 \\
		RAFT\cite{teed2020RAFTRecurrent} & 
		{\color{gray}2.25} &{\color{gray}2.49} & {\color{gray}1.63}& 1.51& 1.52 & 11 \\
		SEA-RAFT\cite{wang2024SEARAFTSimple} & 
		{\color{gray}2.22} &{\color{gray}1.74} & 0.87& \underline{0.77}& \textbf{1.2}& 4\\
		FlowDiffuser\cite{luo2024FlowDiffuserAdvancing} & 
		{\color{gray}2.06} &0.88 & 0.66& 1.25& 1.56 & 6\\
		MemFlow\cite{dong2024MemFlowOptical} & 
		{\color{gray}1.88} &1.13 & 0.78& 1.22& 1.52 & 5\\
		StreamFlow\cite{sun2024StreamFlowStreamlined} & 
		{\color{gray}1.8} &0.74 & 0.55& \textbf{0.72}& \underline{1.32} &\underline{2} \\
		DpFlow\cite{morimitsu2025DPFlowAdaptive} & 
		{\color{gray}1.75} &0.75 & 0.76& 1.25& {\color{gray}1.58} & 8 \\
		vSTMD (Ours) & 
		1.42 & \underline{0.27} & \underline{0.36}& 1.17& 1.57 & 3\\
		vSTMD-F (Ours) & 
		0.88 &\textbf{0.21} & \textbf{0.29}& 0.92& 1.49 &\textbf{1} \\
		\bottomrule
	\end{tabular}
	\label{Tab:orientation}
\end{table}
\endgroup

Table \ref{Tab:orientation} compares the AAE performance across various velocity regimes, where a detailed version is listed in Sect. I-B of  \textbf{Supplementary Materials}. Our proposed vSTMD-F achieves the best average rank (1), demonstrating superior precision, particularly in the medium-velocity ranges ($0.6-3$ pixels/frame) with a minimum AAE of 0.21.

In contrast, traditional STMD models (DSTMD \cite{Wang2020DSTMD}, STMD+ \cite{Wang2020STMDpuls}, and ApgSTMD \cite{Wang2022attention}) are primarily effective for sub-pixel motion ($\le 0.5$ pixels/frame) and fail to provide valid outputs as velocity increases. Meanwhile, mainstream deep learning methods, such as StreamFlow \cite{sun2024StreamFlowStreamlined} and SEA-RAFT \cite{wang2024SEARAFTSimple}, exhibit their strengths in the high-speed regime ($3.1-5$ pixels/frame). When the velocity exceeds the target's physical size ($5$ pixels), which belongs to a large displacement, the angular estimation becomes extremely difficult for all methods, with AAEs exceeding 1 rad.

In conclusion, by effectively bridging the gap between sub-pixel precision and large-displacement stability, vSTMD-F attains the highest overall rank, proving its reliability for real-world applications where target velocities are highly dynamic.

\subsection{Real-World Dataset Comparison}

\begin{table*}[t]
	\centering
	\caption{Results on the real-world RIST dataset \cite{RIST_DataSet} in terms of localization, orientation, and FPS. Deep-learning baselines are included only for orientation comparison, as they do not provide localization outputs. $^{\dagger}$ indicates methods that can localize targets but cannot determine orientation, and $^{\ddagger}$ marks optical-flow-based approaches that estimate orientation only.}
	\begin{tabular}{crrccccccc} 
		\toprule
		&\multirow{2}*{Method} & \multirow{2}*{Venue}
		& \multicolumn{4}{c}{Localization} & Orientation & \multicolumn{2}{c}{FPS} \\	
		\cmidrule(lr){4-7} \cmidrule(lr){8-8} \cmidrule(lr){9-10}
		&& & mAUC(\%)$\uparrow$ & mAR(\%)$\uparrow$ 
		&mAP(\%)$\uparrow$ & mF$_{1}$(\%)$\uparrow$ & AAE(rad)$\downarrow$  & CPU$\uparrow$ & GPU$\uparrow$	\\
		\midrule
		\multirow{9}*{\rotatebox{90}{Natural architecture}}&$^{\dagger}$ESTMD\cite{Wiederman2008ESTMD}  &  Plos-One'08 	
		& 28.30  & 19.50 & 12.17 & 14.93 & - & 64 & - 	\\
		&DSTMD\cite{Wang2020DSTMD}	& T-Cybern'20
		& 28.40 &21.06&13.41&16.35 & 1.02 & 18 & -  \\
		&STMD+\cite{Wang2020STMDpuls}  & T-NNLS'20
		& 37.03 &27.89&18.65&22.31 &0.91 & 11 & - \\
		&$^{\dagger}$LC11\cite{tanaka2020object}& Current Biology'20
		&1.47 &  1.43 &  0.48 &  0.63  & - & 4 & -  \\
		&$^{\dagger}$FeedbackSTMD\cite{Wang2021time} & T-NNLS'21
		& 39.58 &24.77&15.07&18.70 & - & 55 & - \\
		&ApgSTMD\cite{Wang2022attention}   & T-Cybern'22
		& 26.52 & 8.68& 5.75& 6.75 & \underline{0.82} & 6 & - \\
		&$^{\dagger}$LC18\cite{klapoetke2022FunctionallyOrdered}& Neuron'22
		& 1.47 & 0.22 & 0.18 & 0.18  & - & - &  148 \\
		&$^{\dagger}$$\mathcal{F}$STMD\cite{ling2022mathematical}    & Fnbot'23
		& 40.22 &30.95&20.36&24.51 & - & 81 & - \\
		&$^{\dagger}$Frac-STMD\cite{Xu2023frac}& Neurocomputing'23
		& \underline{45.65} & \underline{31.43} & \underline{20.66} & \underline{24.88} & - & \textbf{249} & -  \\
		\cmidrule(lr){1-3}
		\multirow{6}*{\rotatebox{90}{Data-driven}}&$^{\ddagger}$RAFT\cite{teed2020RAFTRecurrent} & ECCV'20
		& - & - & - & - & 1.29 & - & 10  \\
		&$^{\ddagger}$SEA-RAFT\cite{wang2024SEARAFTSimple} &ECCV'24
		& - & - & - & - & 1.25 & - & 44  \\
		&$^{\ddagger}$FlowDiffuser\cite{luo2024FlowDiffuserAdvancing} &ECCV'24
		& - & - & - & - & 1.01& - & 2 \\
		&$^{\ddagger}$MemFlow\cite{dong2024MemFlowOptical} &CVPR'24
		& - & - & - & - & 1.08 & - & 5 \\
		&$^{\ddagger}$StreamFlow\cite{sun2024StreamFlowStreamlined} &NeurIPS'24
		& - & - & - & - & 1.09 & - & 6 \\
		&$^{\ddagger}$DpFlow\cite{morimitsu2025DPFlowAdaptive} & CVPR'25
		& - & - & - & - & \underline{0.74} & - & 6 \\
		\cmidrule(lr){1-3}
		&vSTMD (Ours)& -
		& \underline{47.35} & \underline{38.64} & \underline{28.06} & \underline{32.44} & \underline{0.71} & \underline{168} & \textbf{608}  \\
		&vSTMD-F (Ours)& -
		& \textbf{59.38} & \textbf{46.18} & \textbf{34.46} & \textbf{39.33} & \textbf{0.50} & 161 & \underline{567}  \\
		\bottomrule
	\end{tabular}
	\label{Tab:RIST}
\end{table*}

To evaluate the generalization and robustness, we conducted experiments on the real-world RIST dataset \cite{RIST_DataSet}. The comparative results are summarized in Table \ref{Tab:RIST}. 

\subsubsection{Localization Superiority} The proposed vSTMD achieves marginal improvements ($1.7\%$ mAUC and $7\%$ mAR, mAP, and $mF_{1}$) over the best-performing natural baseline, FracSTMD \cite{Xu2023frac}. In terms of the $mF_{1}$ score, vSTMD-F achieves $39.33\%$, representing a $58.0\%$ relative improvement over Frac-STMD ($24.88\%$).
	
\subsubsection{Orientation Accuracy} The proposed vSTMD (0.71 rad) performs on par with the strongest data-driven competitor, DpFlow \cite{morimitsu2025DPFlowAdaptive} (0.74 rad). Further, the vSTMD-F reduces the Average Angular Error (AAE) to 0.50 rad, which is $32.4\%$ and $39\%$ error reductions compared to the DpFlow, and the best directional natural architecture model, ApgSTMD \cite{Wang2022attention} (0.82 rad), respectively. 
	
\subsubsection{Computational Efficiency} vSTMD and vSTMD-F maintain high real-time efficiency, achieving 608 and 567 FPS on the GPU, respectively, which is orders of magnitude faster than deep-learning models (e.g., 60 times faster than DpFLow \cite{morimitsu2025DPFlowAdaptive} and StreamFlow \cite{sun2024StreamFlowStreamlined}, and 12 times faster SEA-RAFT \cite{wang2024SEARAFTSimple}).

\subsubsection{Visualization} 
\begin{figure*}[!t]
	\centering
	\includegraphics[width=1\textwidth]{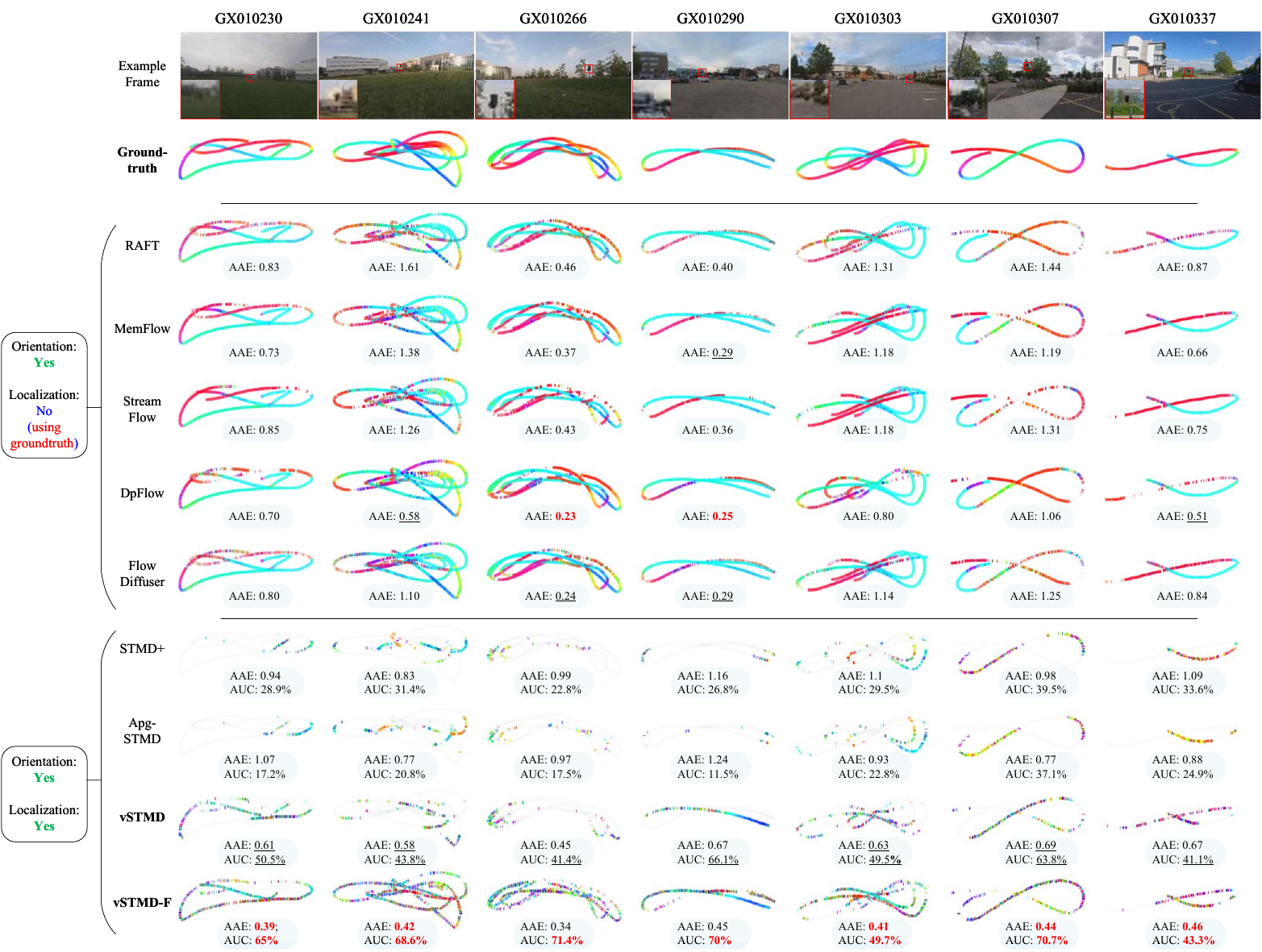}
	\caption{Demonstration of evaluation in RIST \cite{RIST_DataSet} with orientation criteria AAE(rad)$\downarrow$ and localization criteria AUC(\%)$\uparrow$. Notably, the results of optical flow approaches are recorded by the given position ground truth, as those methods are only designed for orientation (including RAFT\cite{teed2020RAFTRecurrent}, MemFlow\cite{dong2024MemFlowOptical}, StreamFlow\cite{sun2024StreamFlowStreamlined}, DpFlow\cite{morimitsu2025DPFlowAdaptive}, and FlowDiffuser\cite{luo2024FlowDiffuserAdvancing}). In contrast, STMD-architectures, such as STMD+\cite{Wang2020STMDpuls}, Apg-STMD\cite{Wang2022attention}, vSTMD (ours), and vSTMD-F (ours), are able to recognize direction and location simultaneously. The first row illustrates scenarios where extremely tiny targets are highlighted by a red box and zoomed in at the bottom left. The second row shows the ground truth with motion direction given by different colors (HSV space) along the motion trajectory.}
	\label{Fig:result_in_RIST}
\end{figure*}

We also visualize part of the videos and models' results in Fig. \ref{Fig:result_in_RIST}, where the direction outputs are represented in HSV space (hue for direction), with the directional metric AAE ($\downarrow$) and the localization metric AUC ($\uparrow$) listed below each result. Notably, the optical flow methods rely on ground-truth positions for evaluation, whereas the STMD architectures jointly predict direction and location. For orientation accuracy, DpFlow \cite{morimitsu2025DPFlowAdaptive} achieves the lowest AAE in GX010266 (0.23) and GX010290 (0.25), while vSTMD-F outperforms in the remaining cases. In terms of AUC, vSTMD-F performs well across all sequences, while vSTMD shows superior results compared to STMD+ \cite{Wang2020STMDpuls} and ApgSTMD \cite{Wang2022attention}.

\subsection{Ablation} \label{Sect_Ablation}

\begingroup
\setlength{\tabcolsep}{5pt} 
\renewcommand{\arraystretch}{1.2} 
\begin{table*}[!t]
	\centering
	\caption{We ablate the GF (Gaussian Filter), the proposed cIDP, and CDGC, as well as the Feedback pathway on the RIST \cite{RIST_DataSet}.}
	\begin{tabular}{c c c cccccc c c c c c} 
		\toprule
		\multicolumn{3}{c}{Description} & \textbf{GF} &  Delay & \textbf{cDIP} & Iso-DGC & \textbf{CDGC} &
		\textbf{Feedback} & \multicolumn{4}{c}{Localization} & Orientation \\
		\cmidrule(lr){10-13} \cmidrule(lr){14-14}
		&&&&&&&&&\parbox{0.9cm}{\centering mAUC\\(\%)$\uparrow$}&
		\parbox{0.9cm}{\centering mAR\\(\%)$\uparrow$}&
		\parbox{0.9cm}{\centering mAP\\(\%)$\uparrow$}&
		\parbox{0.9cm}{\centering mF$_{1}$\\(\%)$\uparrow$}&
		\parbox{0.9cm}{\centering AAE\\(rad)$\downarrow$}
		\\
		\midrule
		\multicolumn{3}{c}{Baseline: \textbf{vSTMD}} &$\checkmark$ & &$\checkmark$ & &$\checkmark$ & & \textbf{47.35} &\textbf{38.64} & \textbf{28.06} & \textbf{32.44}& \textbf{0.71} \\
		\cmidrule(lr){1-3}
		\multirow{4}*{\centering \rotatebox{90}{Ablation}} &  &without GF& &&$\checkmark$ & &$\checkmark$ & & 27.36  &24.13 & 16.07 & 19.24& 1.29   \\
		\cmidrule(lr){2-3}
		& &without cIDP &$\checkmark$ &$\checkmark$&&&$\checkmark$& & 36.00  &26.30 & 16.91 & 20.47& 1.94  \\
		\cmidrule(lr){2-3}
		&\multirow{2}*{\rotatebox{90}{CDGC}} &Iso-DGC&$\checkmark$ &&$\checkmark$&$\checkmark$&& & 47.35  &38.64 & 28.06 & 32.44& 0.75 \\
		&&Hungarian Matching&$\checkmark$ &&$\checkmark$&&& & 47.35  &38.64 & 28.06 & 32.44& 1.60  \\
		\midrule
		\multicolumn{3}{c}{Baseline:\textbf{ vSTMD-F}} &$\checkmark$ & &$\checkmark$ & &$\checkmark$ &$\checkmark$ (no $\tau$) & \textbf{59.38} &\textbf{46.18} & \textbf{34.46} & \textbf{39.33} & \textbf{0.50} \\
		\cmidrule(lr){1-3}
		\multirow{5}*{\centering \rotatebox{90}{Ablation}} &  &without GF& &&$\checkmark$ & &$\checkmark$ &$\checkmark$ (no $\tau$) & 48.18  &33.88 & 23.28 & 27.40& 0.74  \\
		\cmidrule(lr){2-3}
		&\multirow{2}*{\rotatebox{90}{cIDP}} &without cIDP &$\checkmark$ &$\checkmark$&&&$\checkmark$&$\checkmark$ (no $\tau$) & 41.89  &26.47 & 15.19 & 19.23& 1.98 \\
		&&Feedback STMD \cite{Wang2021time} &$\checkmark$ &$\checkmark$&&&&$\checkmark$ ($\tau$ used) & 39.27  &24.34 & 14.82 & 18.39& -  \\
		\cmidrule(lr){2-3}
		&\multirow{2}*{\rotatebox{90}{CDGC}} &Iso-DGC&$\checkmark$ &&$\checkmark$&$\checkmark$&&$\checkmark$ (no $\tau$) & 59.38 &46.18 & 34.46 & 39.33& 0.91  \\
		&&Hungarian Matching&$\checkmark$ &&$\checkmark$&&&$\checkmark$ (no $\tau$) & 59.38 &46.18 & 34.46 & 39.33& 1.57  \\
		\bottomrule
	\end{tabular}
	\label{Tab:ablation}
\end{table*}
\endgroup

To evaluate the individual contributions of the proposed modules, we conducted systematic ablation studies focusing on three core components: the Gaussian Filter (GF), the cross-Inhibition Dynamic Potential (cIDP), and the Collaborative Directional Gradient Calculation (CDGC) mechanism. These evaluations were performed across both the vSTMD and vSTMD-F models. The experimental results are summarized in Table \ref{Tab:ablation}.

\subsubsection{Gaussian Filter}
The absence of the GF component results in significant performance degradation across all metrics. For vSTMD, localization accuracy decreases by over $35\%$, and the orientation error worsens from 0.71 rad to 1.29 rad (an $81.6\%$ increase). 

\subsubsection{cIDP}
The cIDP component yielded consistent enhancements across all localization metrics. Notably, since the CDGC mechanism is functionally coupled with cIDP, the absence of the latter resulted in a substantial increase in orientation error—rising from 0.71 to 1.94 rad for vSTMD and from 0.50 to 1.98 rad for vSTMD-F. 

\subsubsection{CDGC}
The specialized CDGC design achieved superior directional selectivity compared to the Isotropic DGC (Iso-DGC) baseline, resulting in $5\%$ and $80\%$ reductions in orientation error for vSTMD and vSTMD-F, respectively. Furthermore, we evaluated the Hungarian Matching, a representative tracking-after-detection strategy for motion direction calculation. However, this methodology performs poorly in real-time motion detection scenarios, as it relies on stable and continuous target detection. Such requirements are often difficult to satisfy in dynamic environments where detections may be intermittent, whereas our integrated approach maintains robustness.

\subsubsection{Feedback}
The contribution of the feedback loop is quantified by the performance gap between vSTMD and vSTMD-F, where localization performance improves by approximately $25\%$, and the orientation error is reduced by $40\%$. Furthermore, our instantaneous feedback mechanism (without delay $\tau$) performs on par with the Feedback STMD model \cite{Wang2021time} in localization. This suggests that the removal of the temporal delay $\tau$ from the feedback pathway is feasible while maintaining high detection efficacy.

\subsection{Module Time Cost Analysis}
Table \ref{Tab:module_time} summarizes the computational costs of each module. The proposed cIDP-and-correlate achieves a slightly lower latency than the conventional mechanism \cite{Wiederman2008ESTMD}. A more significant improvement is observed for the CDGC module, which reduces the CPU time from 51.78 ms to 6.81 ms. This nearly eight-fold reduction (approximately $13.1\%$) highly aligns with our theoretical analysis in Sect. \ref{Sect:orientation_pipeline}, confirming that the proposed mechanisms substantially alleviate computational overhead.

\begingroup
\renewcommand{\arraystretch}{1.2} 
\begin{table}[t]
	\centering
	\caption{Time-cost comparison of the conventional delay-and-correlate mechanism, Iso-DGC, and the proposed cIDP and CDGC mechanisms.}
	\begin{tabular}{c c c c} 
		\toprule
		& \multirow{2}{*}{Module} & \multicolumn{2}{c}{Time Cost (ms)}\\
		\cmidrule(lr){3-4}
		&& CPU & GPU \\
		\midrule
		\multirow{2}*{Localization}&Delay-and-correlate\cite{Wiederman2008ESTMD}& 2.18 & 1.76 \\
		&\textbf{cIDP-and-correlate}& 2.15 & 1.01 \\
		\cmidrule(lr){1-2}
		\multirow{2}*{Orientation}&Iso-DGC\cite{Wang2020DSTMD}& 51.78 & 27.58 \\
		&\textbf{CDGC}& 6.81 & 4.30 \\
		\bottomrule
	\end{tabular}
	\label{Tab:module_time}
\end{table}
\endgroup

\section{Discussion} \label{Sect:Discussion}

Experimental results indicate that the cIDP module improves various velocity motion localization while the CDGC mechanism facilitates robust direction identification, for ET-targets. The integration of these components yields an advanced STMD model capable of detecting ET-target motion across a wide range of velocities. Although the proposed method demonstrates significant performance gains, potential optimizations and broader applications remain. Future research directions are detailed below.

\subsection{STMD-Related Mechanism Extension}
While we demonstrate vSTMD boosting Feedback STMD \cite{Wang2021time} by about 20\% (vSTMD-F), it is crucial to consider its compatibility with other existing mechanisms such as feedback loops \cite{ling2022mathematical}, contrast pathways \cite{Wang2020STMDpuls}, and attention and prediction mechanisms \cite{Wang2022attention}. Many current models incorporate parameters that are inherently tied to pre-defined velocity assumptions. One important direction for future research is exploring how these mechanisms can be adapted or re-engineered to align with the new backbone, which operates without needing prior velocity-related parameters. 

\subsection{Learning-Based Extension}

Although vSTMD is intentionally designed as a natural architecture–based and learning-free model, its structure naturally suggests a pathway toward hybrid learning. In particular, the proposed cIDP mechanism, while functioning as an interpretable implementation of cross-inhibition between opposite-polarity channels, can also be viewed as an initial abstraction of the insect medulla.

However, neurophysiological studies indicate that the medulla contains at least ten distinct strata \cite{Arenz2017temporal, takemura2013visual, fisher2015class}, each contributing to progressively refined feature extraction. This biological insight implies that the current fixed-parameter cIDP formulation could in the future be replaced or augmented by a shallow learnable module (e.g., a lightweight 10-layer network) that mirrors the medulla’s laminar organization. Such a hybrid module would retain the architectural interpretability while granting the system the flexibility to adapt to diverse background structures, dynamic illumination changes, or cluttered environments.

\subsection{Task Extension}

As a low-level visual task, ET-target motion detection can be combined with the following high-level tasks and real applications.

\subsubsection{Search and Rescue Task}

Video \textbf{supplemental materials} demonstrating maritime search-and-rescue scenarios are provided based on the SeaDronesSee\footnote{\url{https://seadronessee.cs.uni-tuebingen.de/home}} dataset \cite{varga2022seadronessee}. Although these demonstrations were executed on an RTX 3050 laptop GPU, they highlight a fundamental architectural advantage of the proposed model. Unlike conventional UAV workflows that compromise detection accuracy via 4K downsampling, our lightweight model processes raw imagery directly with minimal overhead, preserving critical spatial details to reduce missed detections and accelerate response times.

Future work will explore the integration of this model with a learning-based human-detection head, where our method serves as a lightweight pre-filter to generate region proposals for downstream deep neural networks. This approach is expected to further optimize UAV search missions by significantly reducing the total computational load while maintaining high detection sensitivity\footnote{\url{https://p2code-project.eu/}}.

\subsubsection{Moving ET-target Detection}

To further investigate the potential of the proposed model in a heterogeneous task environment, we provide a cross-task comparison on the XS-VID dataset \cite{guo2024xs}. Unlike conventional deep learning (DL) detectors that rely heavily on category-specific semantic features, our model focuses on class-agnostic motion detection. Therefore, we adopt Average Recall (AR) as the primary metric to evaluate the quality of motion-based proposals.

Table \ref{Tab:object_detection} highlights the superior performance-efficiency trade-off of vSTMD-F. While the yoloft-L \cite{guo2024xs} suffers a performance collapse on ET targets ($6.55\%$ AR), our model maintains a robust $52.86\%$. Compared to the bio-inspired FracSTMD \cite{Xu2023frac}, vSTMD-F achieves a significant recall gain ($+22.23\%$ for ET targets) while sustaining a high-speed inference of 132.9 FPS. Although yoloft-L provides semantic discrimination, its low frame rate (7.9 FPS) is prohibitive for real-time tasks. Consequently, vSTMD-F serves as an ideal high-speed motion proposal generator, bridging the gap between low-level motion perception and high-level semantic analysis in resource-constrained scenarios.

\begingroup
\renewcommand{\arraystretch}{1.2} 
\begin{table}[t]
	\centering
	\caption{Comparison of detection performance (AR, \%) and speed (FPS) on the XS-VID dataset \cite{guo2024xs}. ``All'' and ``Move'' denote the Average Recall for all targets and moving targets, respectively.}
	\begin{tabular}{cccccc} 
		\toprule
		\multirow{3}{*}{Model}
		& \multicolumn{2}{c}{\parbox{2.2cm}{\centering Extremely Small \\($\le 12\times12$ pixels)}}
		& \multicolumn{2}{c}{\parbox{2.2cm}{\centering Extremely Tiny \\($\le 8\times8$ pixels)}}
		& \multirow{3}{*}{\parbox{0.8cm}{\centering FPS \\(GPU)}}\\
		\cmidrule(lr){2-3} \cmidrule(lr){4-5}
		& All & Move & All & Move & \\
		\midrule
		yoloft-L \cite{guo2024xs} & \underline{52.37} & \underline{65.93} & 6.55 & 7.38 & 7.9 \\
		FracSTMD \cite{Xu2023frac} & 20.63 & 21.81 & \underline{30.63} & \underline{33.56} & \textbf{213.5} \\
		\textbf{vSTMD-F} & \textbf{62.95} & \textbf{67.09} & \textbf{52.86} & \textbf{54.39} & \underline{132.9} \\
		\bottomrule
	\end{tabular}
	\label{Tab:object_detection}
\end{table}
\endgroup

Furthermore, the inherent motion-centric characteristics of the proposed model facilitate its deployment on dynamic platforms, such as satellites, by integrating with advanced remote sensing frameworks \cite{gao2023attention, gao2023task}. Additionally, the model is well-suited for reformulation into a Spiking Neural Network (SNN) architecture to seamless integration with neuromorphic perception systems \cite{li2025brain}, which process asynchronous event streams. Such a paradigm shift from frame-based to event-based processing offers the potential for significantly reduced power consumption and sub-millisecond temporal resolution in complex detection tasks.

\section{Conclusion} \label{Sect:Conclusion}

This paper presents the vSTMD and vSTMD-F frameworks, representing a significant advancement in bio-inspired motion detection for extremely tiny (ET-) targets at various velocities. By introducing the cross-Inhibition Dynamic Potential (cIDP) and the first Collaborative Directional Gradient Calculation (CDGC), we have addressed the long-standing multi-velocity detection bottleneck that limited previous STMD-based models. Results on RIST demonstrate that the vSTMD establishes a robust baseline for joint localization and orientation. Moreover, the feedback-facilitated variant, vSTMD-F, achieves a $58\%$ $mF_{1}$ improvement (localization) and a $39\%$ reduction in AAE (Average Angular Error, orientation) compared to current state-of-the-art methods. Despite these gains in accuracy, both models preserve their inherent computational efficiency, operating at over 560 FPS. The robustness and generalization of our approach are further validated through cross-task evaluations and maritime search-and-rescue applications. Ultimately, this work establishes a high-performance baseline for real-time ET-target motion detection, paving the way for future research into hybrid bio-inspired learning mechanisms and complex vision task extensions. 

\bibliography{vSTMD_bibfile}

@article{guo2024xs,
  title={XS-VID: An Extremely Small Video Object Detection Dataset},
  author={Guo, Jiahao and Xu, Ziyang and Wu, Lianjun and Gao, Fei and Liu, Wenyu and Wang, Xinggang},
  journal={arXiv preprint arXiv:2407.18137},
  year={2024}
}

@inproceedings{li2025brain,
	title={Brain-Inspired Spiking Neural Networks for Energy-Efficient Object Detection},
	author={Li, Ziqi and Gao, Tao and An, Yisheng and Chen, Ting and Zhang, Jing and Wen, Yuanbo and Liu, Mengkun and Zhang, Qianxi},
	booktitle={Proceedings of the Computer Vision and Pattern Recognition Conference},
	pages={3552--3562},
	year={2025}
}

@article{gao2023attention,
	title={Attention-free global multiscale fusion network for remote sensing object detection},
	author={Gao, Tao and Li, Ziqi and Wen, Yuanbo and Chen, Ting and Niu, Qianqian and Liu, Zixiang},
	journal={IEEE transactions on geoscience and remote sensing},
	volume={62},
	pages={1--14},
	year={2023},
	publisher={IEEE}
}

@article{gao2023task,
	title={A task-balanced multiscale adaptive fusion network for object detection in remote sensing images},
	author={Gao, Tao and Liu, Zixiang and Zhang, Jing and Wu, Guiping and Chen, Ting},
	journal={IEEE Transactions on Geoscience and Remote Sensing},
	volume={61},
	pages={1--15},
	year={2023},
	publisher={IEEE}
}

@inproceedings{varga2022seadronessee,
	title={Seadronessee: A maritime benchmark for detecting humans in open water},
	author={Varga, Leon Amadeus and Kiefer, Benjamin and Messmer, Martin and Zell, Andreas},
	booktitle={Proceedings of the IEEE/CVF Winter Conference on Applications of Computer Vision},
	pages={2260--2270},
	year={2022} }

@article{yasir2021review,
  title={Review on real time background extraction: Models, applications, environments, challenges and evaluation approaches},
  author={Yasir, Maryam and Ali, Yossra},
  year={2021},
  publisher={International Association of Online Engineering}
}

@article{kim2022learning,
  title={Learning open-world object proposals without learning to classify},
  author={Kim, Dahun and Lin, Tsung-Yi and Angelova, Anelia and Kweon, In So and Kuo, Weicheng},
  journal={IEEE Robotics and Automation Letters},
  volume={7},
  number={2},
  pages={5453--5460},
  year={2022},
  publisher={IEEE}
}

@article{alfarano2024EstimatingOptical,
  title = {Estimating Optical Flow: {{A}} Comprehensive Review of the State of the Art},
  shorttitle = {Estimating Optical Flow},
  author = {Alfarano, Andrea and Maiano, Luca and Papa, Lorenzo and Amerini, Irene},
  year = {2024},
  month = dec,
  journal = {Computer Vision and Image Understanding},
  volume = {249},
  pages = {104160},
  issn = {10773142},
  doi = {10.1016/j.cviu.2024.104160},
  urldate = {2024-12-01}
}

@inproceedings{lucas1981IterativeImage,
  title = {An {{Iterative Image Registration Technique}} with an {{Application}} to {{Stereo Vision}}},
  booktitle = {{{IJCAI}}'81: 7th International Joint Conference on {{Artificial}} Intelligence},
  author = {Lucas, Bruce D and Kanade, Takeo},
  year = {1981},
  month = aug,
  volume = {2},
  pages = {674--679},
  address = {Vancouver, Canada},
  urldate = {2025-05-20}
}

@inproceedings{dong2024MemFlowOptical,
  title = {{{MemFlow}}: {{Optical Flow Estimation}} and {{Prediction}} with {{Memory}}},
  shorttitle = {{{MemFlow}}},
  booktitle = {2024 {{IEEE}}/{{CVF Conference}} on {{Computer Vision}} and {{Pattern Recognition}} ({{CVPR}})},
  author = {Dong, Qiaole and Fu, Yanwei},
  year = {2024},
  month = jun,
  pages = {19068--19078},
  publisher = {IEEE},
  address = {Seattle, WA, USA},
  doi = {10.1109/CVPR52733.2024.01804},
  urldate = {2025-05-19},
  copyright = {https://doi.org/10.15223/policy-029},
  isbn = {979-8-3503-5300-6}
}

@inproceedings{luo2024FlowDiffuserAdvancing,
  title = {{{FlowDiffuser}}: {{Advancing Optical Flow Estimation}} with {{Diffusion Models}}},
  shorttitle = {{{FlowDiffuser}}},
  booktitle = {2024 {{IEEE}}/{{CVF Conference}} on {{Computer Vision}} and {{Pattern Recognition}} ({{CVPR}})},
  author = {Luo, Ao and Li, Xin and Yang, Fan and Liu, Jiangyu and Fan, Haoqiang and Liu, Shuaicheng},
  year = {2024},
  month = jun,
  pages = {19167--19176},
  publisher = {IEEE},
  address = {Seattle, WA, USA},
  doi = {10.1109/CVPR52733.2024.01813},
  urldate = {2025-05-19},
  copyright = {https://doi.org/10.15223/policy-029},
  isbn = {979-8-3503-5300-6}
}

@inproceedings{morimitsu2025DPFlowAdaptive,
  title={DPFlow: Adaptive Optical Flow Estimation with a Dual-Pyramid Framework},
  author={Morimitsu, Henrique and Zhu, Xiaobin and Cesar, Roberto M and Ji, Xiangyang and Yin, Xu-Cheng},
  booktitle={Proceedings of the Computer Vision and Pattern Recognition Conference},
  pages={17810--17820},
  year={2025}
}

@inproceedings{sun2024StreamFlowStreamlined,
  title = {{{StreamFlow}}: {{Streamlined Multi-Frame Optical Flow Estimation}} for {{Video Sequences}}},
  booktitle = {38th {{Conference}} on {{Neural Information Processing Systems}} ({{NeurIPS}} 2024).},
  author = {Sun, Shangkun and Liu, Jiaming and Li, Huaxia and Liu, Guoqing and Li, Thomas H and Gao, Wei},
  year = {2024}
}

@inproceedings{teed2020RAFTRecurrent,
  title={Raft: Recurrent all-pairs field transforms for optical flow},
  author={Teed, Zachary and Deng, Jia},
  booktitle={European conference on computer vision},
  pages={402--419},
  year={2020},
  organization={Springer}
}

@inproceedings{wang2024SEARAFTSimple,
  title={Sea-raft: Simple, efficient, accurate raft for optical flow},
  author={Wang, Yihan and Lipson, Lahav and Deng, Jia},
  booktitle={European Conference on Computer Vision},
  pages={36--54},
  year={2024},
  organization={Springer}
}

@article{wang2024BioInspiredSmall,
  title = {Bio-{{Inspired Small Target Motion Detection With Spatio-Temporal Feedback}} in {{Natural Scenes}}},
  author = {Wang, Hongxin and Zhong, Zhiyan and Lei, Fang and Peng, Jigen and Yue, Shigang},
  year = {2024},
  journal = {IEEE Transactions on Image Processing},
  volume = {33},
  pages = {451--465},
  issn = {1057-7149, 1941-0042},
  doi = {10.1109/TIP.2023.3345153},
  urldate = {2024-04-12},
  copyright = {https://ieeexplore.ieee.org/Xplorehelp/downloads/license-information/IEEE.html}
}

@article{klapoetke2022FunctionallyOrdered,
  title = {A Functionally Ordered Visual Feature Map in the {{Drosophila}} Brain},
  author = {Klapoetke, Nathan C. and Nern, Aljoscha and Rogers, Edward M. and Rubin, Gerald M. and Reiser, Michael B. and Card, Gwyneth M.},
  year = {2022},
  month = may,
  journal = {Neuron},
  volume = {110},
  number = {10},
  pages = {1700-1711.e6},
  issn = {08966273},
  doi = {10.1016/j.neuron.2022.02.013},
  urldate = {2024-11-18}
}

@article{tanaka2020object,
  title={Object-displacement-sensitive visual neurons drive freezing in Drosophila},
  author={Tanaka, Ryosuke and Clark, Damon A},
  journal={Current Biology},
  volume={30},
  number={13},
  pages={2532--2550},
  year={2020},
  publisher={Elsevier}
}

@article{geurten2007NeuralMechanisms,
  title = {Neural Mechanisms Underlying Target Detection in a Dragonfly Centrifugal Neuron},
  author = {Geurten, Bart and Nordstr{\"o}m, Karin and Sprayberry, Jordanna and Bolzon, Douglas and O'Carroll, David},
  year = {2007},
  month = oct,
  journal = {The Journal of experimental biology},
  volume = {210},
  pages = {3277--84},
  doi = {10.1242/jeb.008425}
}

@article{zhai2021OpticalFlow,
  title = {Optical Flow and Scene Flow Estimation: {{A}} Survey},
  shorttitle = {Optical Flow and Scene Flow Estimation},
  author = {Zhai, Mingliang and Xiang, Xuezhi and Lv, Ning and Kong, Xiangdong},
  year = {2021},
  month = jun,
  journal = {Pattern Recognition},
  volume = {114},
  pages = {107861},
  issn = {00313203},
  doi = {10.1016/j.patcog.2021.107861},
  urldate = {2025-05-19}
}

@article{ying2025VisibleThermalTiny,
  title = {Visible-{{Thermal Tiny Object Detection}}: {{A Benchmark Dataset}} and {{Baselines}}},
  shorttitle = {Visible-{{Thermal Tiny Object Detection}}},
  author = {Ying, Xinyi and Xiao, Chao and An, Wei and Li, Ruojing and He, Xu and Li, Boyang and Cao, Xu and Li, Zhaoxu and Wang, Yingqian and Hu, Mingyuan and Xu, Qingyu and Lin, Zaiping and Li, Miao and Zhou, Shilin and Sheng, Weidong and Liu, Li},
  year = {2025},
  journal = {IEEE Transactions on Pattern Analysis and Machine Intelligence},
  pages = {1--8},
  issn = {1939-3539},
  doi = {10.1109/TPAMI.2025.3544621},
  urldate = {2025-05-17}
}

@article{tanaka2022neural,
	title={Neural mechanisms to exploit positional geometry for collision avoidance},
	author={Tanaka, Ryosuke and Clark, Damon A},
	journal={Current Biology},
	volume={32},
	number={11},
	pages={2357--2374},
	year={2022},
	publisher={Elsevier}
}

@article{haag1997encoding,
	title={Encoding of visual motion information and reliability in spiking and graded potential neurons},
	author={Haag, Juergen and Borst, Alexander},
	journal={Journal of Neuroscience},
	volume={17},
	number={12},
	pages={4809--4819},
	year={1997},
	publisher={Soc Neuroscience}
}

@article{simmons1999performance,
	title={The performance of synapses that convey discrete graded potentials in an insect visual pathway},
	author={Simmons, Peter J},
	journal={Journal of Neuroscience},
	volume={19},
	number={23},
	pages={10584--10594},
	year={1999},
	publisher={Soc Neuroscience}
}

@article{hengstenberg1977spike,
	title={Spike responses of ‘non-spiking’visual interneurone},
	author={Hengstenberg, Roland},
	journal={Nature},
	volume={270},
	number={5635},
	pages={338--340},
	year={1977},
	publisher={Nature Publishing Group UK London}
}

@article{zettler1973active,
	title={Active and passive axonal propagation of non-spike signals in the retina of Calliphora},
	author={Zettler, F and J{\"a}rvilehto, M},
	journal={Journal of comparative physiology},
	volume={85},
	pages={89--104},
	year={1973},
	publisher={Springer}
}

@article{hodgkin1952quantitative,
	title={A quantitative description of membrane current and its application to conduction and excitation in nerve},
	author={Hodgkin, Alan L and Huxley, Andrew F},
	journal={The Journal of physiology},
	volume={117},
	number={4},
	pages={500},
	year={1952},
	publisher={Wiley}
}

@article{chen2025rigid,
	title={Rigid propagation of visual motion in the insect’s neural system},
	author={Chen, Hao and Fan, Boquan and Li, Haiyang and Peng, Jigen},
	journal={Neural Networks},
	volume={181},
	pages={106874},
	year={2025},
	publisher={Elsevier}
}

@article{chen2024unveiling,
	title={Unveiling the power of Haar frequency domain: Advancing small target motion detection in dim light},
	author={Chen, Hao and Sun, Xuelong and Hu, Cheng and Wang, Hongxin and Peng, Jigen},
	journal={Applied Soft Computing},
	volume={167},
	pages={112281},
	year={2024},
	publisher={Elsevier}
}

@article{wang2023bio,
	title={Bio-Inspired Small Target Motion Detection With Spatio-Temporal Feedback in Natural Scenes},
	author={Wang, Hongxin and Zhong, Zhiyan and Lei, Fang and Peng, Jigen and Yue, Shigang},
	journal={IEEE Transactions on Image Processing},
	year={2023},
	publisher={IEEE}
}

@article{klapoetke2022functionally,
	title={A functionally ordered visual feature map in the Drosophila brain},
	author={Klapoetke, Nathan C and Nern, Aljoscha and Rogers, Edward M and Rubin, Gerald M and Reiser, Michael B and Card, Gwyneth M},
	journal={Neuron},
	volume={110},
	number={10},
	pages={1700--1711},
	year={2022},
	publisher={Elsevier}
}

@article{Arenz2017temporal,
	title={The temporal tuning of the Drosophila motion detectors is determined by the dynamics of their input elements},
	author={Arenz, Alexander and Drews, Michael S and Richter, Florian G and Ammer, Georg and Borst, Alexander},
	journal={Current Biology},
	volume={27},
	number={7},
	pages={929--944},
	year={2017},
	publisher={Elsevier}
}

@article{HR1956EMD,
	url = {https://doi.org/10.1515/znb-1956-9-1004},
	title = {Systemtheoretische Analyse der Zeit-, Reihenfolgen- und Vorzeichenauswertung bei der Bewegungsperzeption des Rüsselkäfers Chlorophanus},
	author = {B. Hassenstein and W. Reichardt},
	pages = {513--524},
	volume = {11},
	number = {9-10},
	journal = {Zeitschrift für Naturforschung B},
	doi = {doi:10.1515/znb-1956-9-1004},
	year = {1956},
	lastchecked = {2024-09-02}
}

@article{BL1965EMD,
	title={The mechanism of directionally selective units in rabbit's retina.},
	author={Barlow, HB and Levick, William R},
	journal={The Journal of physiology},
	volume={178},
	number={3},
	pages={477},
	year={1965},
	publisher={Wiley}
}

@article{frye2015elementary,
	title={Elementary motion detectors},
	author={Frye, Mark},
	journal={Current Biology},
	volume={25},
	number={6},
	pages={R215--R217},
	year={2015},
	publisher={Elsevier}
}

@article{Xu2023frac,
title={A fractional-order visual neural model for small target motion detection},
author={Xu, Mingshuo and Wang, Hongxin and Chen, Hao and Li, Haiyang and Peng, Jigen},
journal={Neurocomputing},
volume={550},
pages={126459},
year={2023},
publisher={Elsevier}
}

@article{Nordstrom2006small,
	title={Small object detection neurons in female hoverflies},
	author={Nordstr{\"o}m, Karin and O'Carroll, David C},
	journal={Proceedings of the Royal Society B: Biological Sciences},
	volume={273},
	number={1591},
	pages={1211--1216},
	year={2006},
	publisher={The Royal Society London}
}

@article{Nordstrom2006insect,
	title={Insect detection of small targets moving in visual clutter},
	author={Nordstr{\"o}m, Karin and Barnett, Paul D and O'Carroll, David C},
	journal={PLoS biology},
	volume={4},
	number={3},
	pages={e54},
	year={2006},
	publisher={Public Library of Science San Francisco, USA}
}

@article{Barnett2007retinotopic,
	title={Retinotopic organization of small-field-target-detecting neurons in the insect visual system},
	author={Barnett, Paul D and Nordstr{\"o}m, Karin and O'carroll, David C},
	journal={Current Biology},
	volume={17},
	number={7},
	pages={569--578},
	year={2007},
	publisher={Elsevier}
}

@article{Nordstrom2012neural,
	title={Neural specializations for small target detection in insects},
	author={Nordstr{\"o}m, Karin},
	journal={Current opinion in neurobiology},
	volume={22},
	number={2},
	pages={272--278},
	year={2012},
	publisher={Elsevier}
}

@article{Fu2019review,
	title={Towards computational models and applications of insect visual systems for motion perception: A review},
	author={Fu, Qinbing and Wang, Hongxin and Hu, Cheng and Yue, Shigang},
	journal={Artificial life},
	volume={25},
	number={3},
	pages={263--311},
	year={2019},
	publisher={MIT Press One Rogers Street, Cambridge, MA 02142-1209, USA journals-info~…}
}

@article{fu2023onoff,
	title={Motion perception based on on/off channels: a survey},
	author={Fu, Qinbing},
	journal={Neural Networks},
	volume={165},
	pages={1--18},
	year={2023},
	publisher={Elsevier}
}

@article{fisher2015class,
	title={A class of visual neurons with wide-field properties is required for local motion detection},
	author={Fisher, Yvette E and Leong, Jonathan CS and Sporar, Katja and Ketkar, Madhura D and Gohl, Daryl M and Clandinin, Thomas R and Silies, Marion},
	journal={Current Biology},
	volume={25},
	number={24},
	pages={3178--3189},
	year={2015},
	publisher={Elsevier}
}

@article{takemura2013visual,
	title={A visual motion detection circuit suggested by Drosophila connectomics},
	author={Takemura, Shin-ya and Bharioke, Arjun and Lu, Zhiyuan and Nern, Aljoscha and Vitaladevuni, Shiv and Rivlin, Patricia K and Katz, William T and Olbris, Donald J and Plaza, Stephen M and Winston, Philip and others},
	journal={Nature},
	volume={500},
	number={7461},
	pages={175--181},
	year={2013},
	publisher={Nature Publishing Group UK London}
}

@article{ling2022mathematical,
	title={Mathematical study of neural feedback roles in small target motion detection},
	author={Ling, Jun and Wang, Hongxin and Xu, Mingshuo and Chen, Hao and Li, Haiyang and Peng, Jigen},
	journal={Frontiers in Neurorobotics},
	volume={16},
	pages={984430},
	year={2022},
	publisher={Frontiers Media SA}
}

@misc{RIST_DataSet,
	author       = {Wang, Hongxin},
	title        = {RIST Data Set.},
	year         = {2019},
	url          = {https://sites.google.com/view/hongxinwang-personalsite/download},
	note         = {Accessed: 2022-01-11}
}

@misc{STMDgit,
	author       = {Xu, Mingshuo},
	title        = {Small-Target-Motion-Detectors, Version 2},
	year         = {2024},
	url          = {https://github.com/MingshuoXu/Small-Target-Motion-Detectors},
	note         = {Accessed: 2024-09-26}
}

@article{Wiederman2008ESTMD,
	title={A Model for the Detection of Moving Targets in Visual Clutter Inspired by Insect Physiology},
	author={ Wiederman, S. D  and  Shoemarker, P. A  and  O'Carroll, D. C },
	journal={PLoS ONE},
	volume={3},
	number={7},
	pages={e2784-},
	year={2008}
}

@article{Wiedermann2013biologically,
	title={Biologically inspired feature detection using cascaded correlations of off and on channels},
	author={Wiedermann, SD and O’Carroll, David C},
	journal={Journal of Artificial Intelligence and Soft Computing Research},
	volume={3},
	year={2013}
}

@article{Wang2020DSTMD,
	title={A Directionally Selective Small Target Motion Detecting Visual Neural Network in Cluttered Backgrounds}, 
	author={Wang, Hongxin and Peng, Jigen and Yue, Shigang},
	journal={IEEE Transactions on Cybernetics}, 
	year={2020},
	volume={50},
	number={4},
	pages={1541-1555}
}

@ARTICLE{Wang2020STMDpuls,  
	author={Wang, Hongxin and Peng, Jigen and Zheng, Xuqiang and Yue, Shigang}, 
	journal={IEEE Transactions on Neural Networks and Learning Systems},   
	title={A Robust Visual System for Small Target Motion Detection Against Cluttered Moving Backgrounds},   
	year={2020},  
	volume={31},  
	number={3},  
	pages={839-853}
}

@article{Wang2022attention,
	title={Attention and Prediction-Guided Motion Detection for Low-Contrast Small Moving Targets},
	author={Wang, Hongxin and Zhao, Jiannan and Wang, Huatian and Hu, Cheng and Peng, Jigen and Yue, Shigang},
	journal={IEEE Transactions on Cybernetics},
	year={2022},
	publisher={IEEE}
}

@article{Wang2021time,
	title={A Time-Delay Feedback Neural Network for Discriminating Small, Fast-Moving Targets in Complex Dynamic Environments},
	author={Wang, Hongxin and Wang, Huatian and Zhao, Jiannan and Hu, Cheng and Peng, Jigen and Yue, Shigang},
	journal={IEEE Transactions on Neural Networks and Learning Systems},
	year={2021},
	publisher={IEEE}
}

@article{Straw2008vision,
	title={Vision egg: an open-source library for realtime visual stimulus generation},
	author={Straw, Andrew D},
	journal={Frontiers in neuroinformatics},
	pages={4},
	year={2008},
	publisher={Frontiers}
}

\end{document}